\documentclass[11pt, a4paper, onecolumn, copyright, goog]{google}

\usepackage[authoryear, sort&compress, round]{natbib}
\usepackage{graphicx}
\usepackage{hyperref}
\usepackage{url}
\usepackage{graphicx} 
\usepackage{float} 
\usepackage{xspace}
\usepackage{subcaption}
\usepackage{booktabs}
\usepackage{multirow}
\usepackage{authblk}
\usepackage{enumitem}
\usepackage{makecell}
\usepackage{pifont}
\usepackage{tcolorbox}
\tcbuselibrary{most}
\usepackage{amssymb}
\usepackage{utfsym}
\usepackage[table]{xcolor}

\definecolor{cellorange}{HTML}{FFE5CC}
\definecolor{cellred}{HTML}{FFC7CE}
\definecolor{cellblue}{HTML}{BDE0FE}
\definecolor{celldarkorange}{HTML}{FFB380}
\definecolor{celldarkred}{HTML}{FF8C8C}

\newcommand{\ourbench}{LogiScope-VQA\xspace}

\newcommand{\obg}[1]{\cellcolor{cellorange}#1}
\newcommand{\rbg}[1]{\cellcolor{cellred}#1}
\newcommand{\bbg}[1]{\cellcolor{cellblue}#1}

\definecolor{propcolor}{HTML}{7B2D8E}  
\definecolor{opencolor}{HTML}{2171B5}  

\newcommand{\ghead}[1]{%
  \makebox[0pt][c]{\small\shortstack{#1}}%
}
\newcommand{\icon}[1]{%
  \raisebox{-0.2em}{\includegraphics[height=0.9em]{#1}}%
}

\definecolor{ForestGreen}{RGB}{0,235,110}
\definecolor{backpurple}{HTML}{E8DEFA}
\definecolor{backyellow}{HTML}{fff0bf}

\keywords{Industrial Benchmark, Logistics Hazard Identification, Visual Question Answering}

\uselogo{} 

\title{\ourbench \raisebox{-0.3cm}{\includegraphics[width=1.2cm, height=1.2cm]{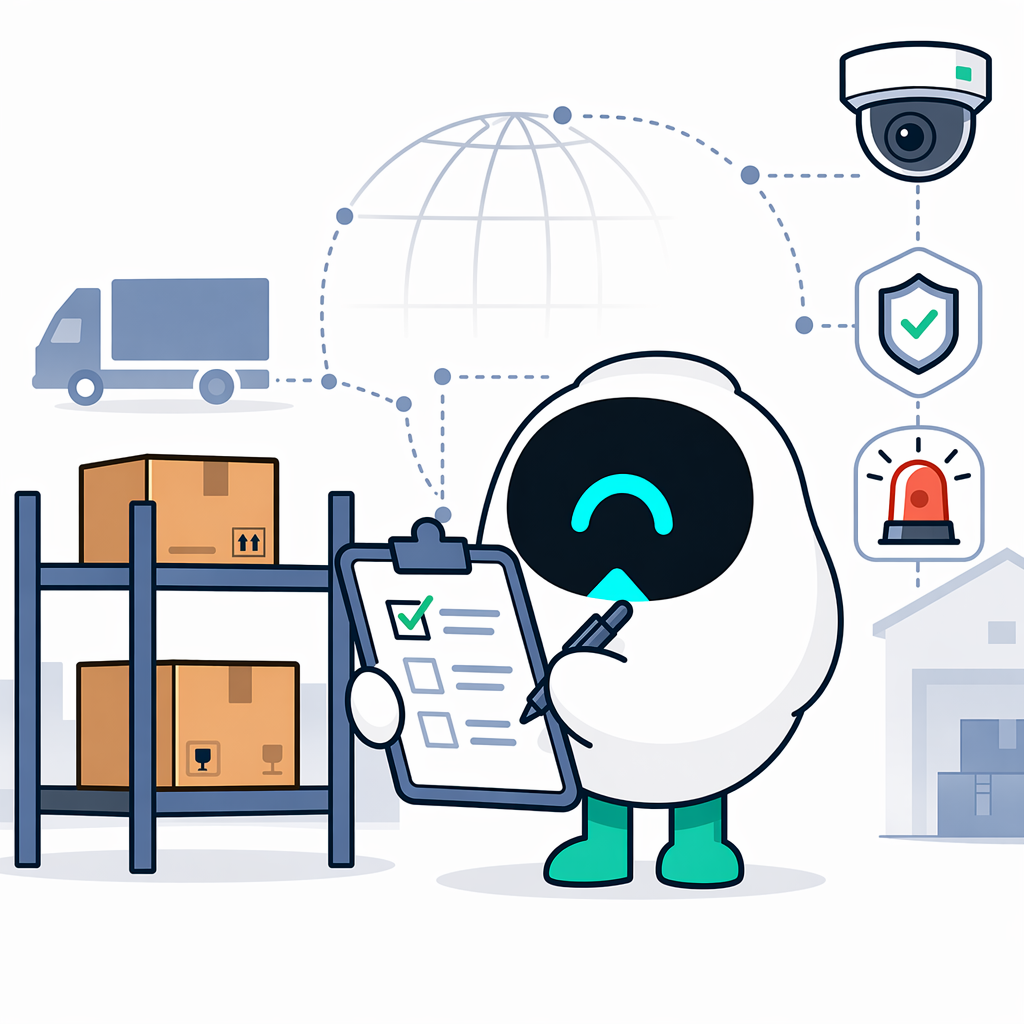}}:
Benchmarking Vision-Language Models for Logistics Hazard Identification in Industrial Scenarios}
\newcommand{\runningtitle}{\ourbench: Benchmarking VLMs for Logistics Hazard Identification in Industrial Scenarios}
\AtBeginDocument{\renewcommand{\today}{}}


\reportnumber{} 

\author[1,*]{Hanjing Zhou}
\author[2,*]{Mingze Yin}
\author[1]{Ying Lian}
\author[1]{Jun Ma}
\author[2,$\dagger$]{Chang‑Yu Hsieh}
\author[1,$\dagger$]{Yanbing Zhou}

\affil[1]{Cainiao Group, Alibaba Group}
\affil[2]{Zhejiang University} 
\affil[*]{\parbox[t]{0.9\linewidth}{Equal contributions. E-mail: \texttt{\{zhj85393, mzyin256\}@gmail.com}}}
\affil[$\dagger$]{\parbox[t]{0.9\linewidth}{Corresponding authors. E-mail: \texttt{kimhsieh@zju.edu.cn, tonychou.zyb@cainiao.com}}}

\begin{abstract}
Large Multimodal Models (LMMs) large-scale deployment in industrial warehouse settings specifically necessitates that models exhibit human-expert-level hazard-oriented perception, understanding, and reasoning capabilities.  
However, the scarcity of real industrial data, tightly coupled to commercial terms, significantly hampers further advancement.
To bridge this gap, we curate \textbf{\ourbench} to investigate the practical applicability of mainstream LMMs in real-world logistics operations.
\ourbench comprises 2,476 images and 2,918 videos primarily sourced from real-world logistics parks, along with 10,274 VQAs meticulously curated and validated by human annotators.
Grounded in 18 core objects and 20 risk types, we devise 39 subtasks aligned with three principal themes: industrial element perception, warehouse knowledge understanding, and potential risk reasoning.
Furthermore, we incorporate dynamic thinking-budget configurations and dual-dimensional risk bias analyses to elucidate the properties of LMMs.
Extensive experiments unveil that even powerful proprietary models, including GPT-5.5, Gemini-3.1-Pro, and Claude-Opus-4.7, exhibit a significant gap relative to human performance.
The unique challenge of jointly integrating perception, understanding, and reasoning for hazard identification poses substantial headroom for further improvement on \ourbench.
We additionally reveal the pervasive security bias issue that impedes LLMs' practical deployment in real-world settings.
The \href{https://github.com/diaoshaoyou/LogiScope-VQA-benchmark}{industrial dataset} is publicly available under the CC BY-NC-SA 4.0 license.
\end{abstract}

\begin{document}

\maketitle

\begin{figure*}[!ht]
    \centering
    \includegraphics[width=1.0\textwidth]{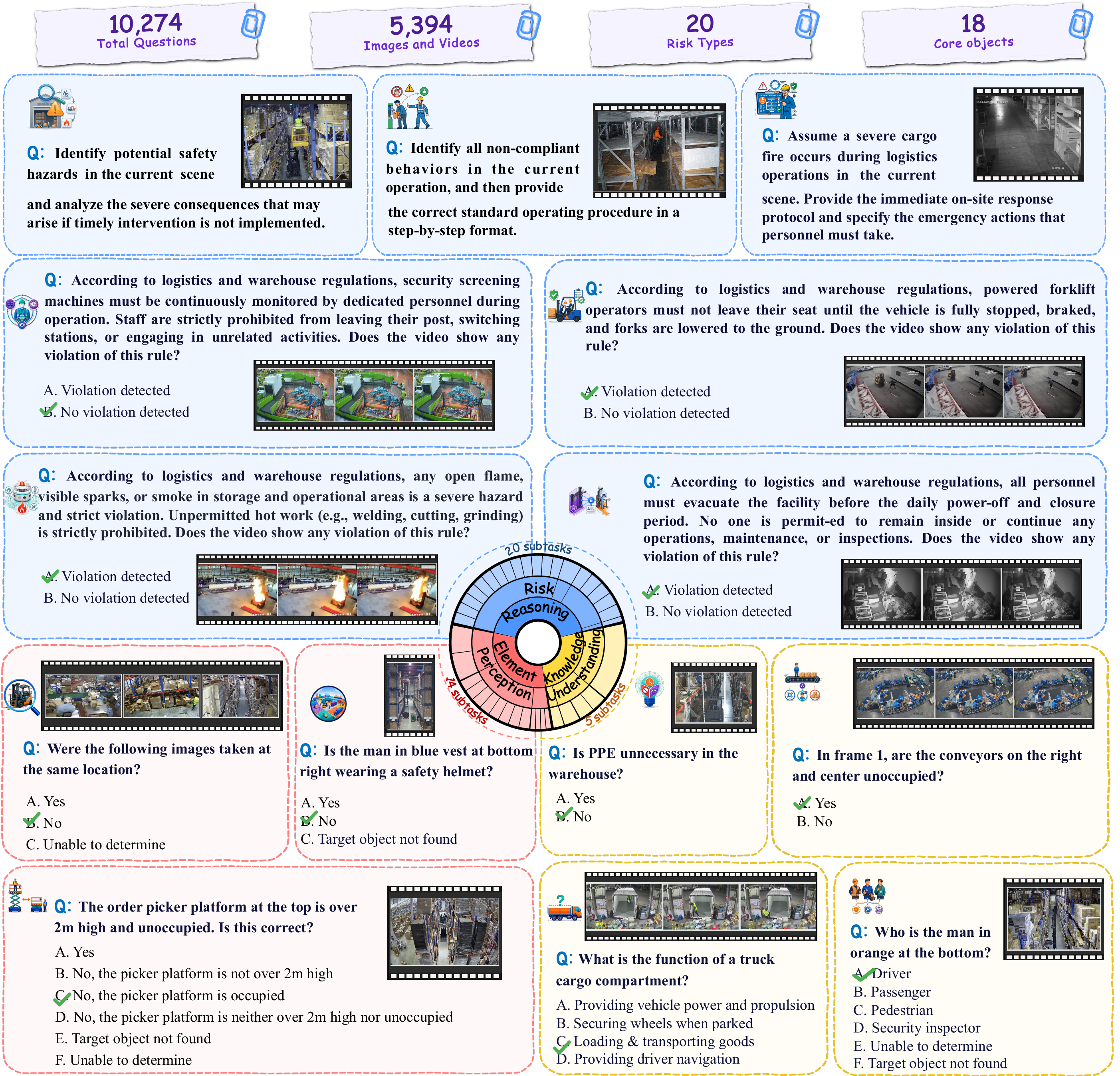}
    \caption{Overview diagram the \ourbench, evaluating LMMs in real-world industrial scenarios across three key competency dimensions: industrial element perception, warehouse knowledge understanding, and potential risk reasoning.
    (The full task taxonomy is provided in Appendix~\ref{task_taxonomy}.)}
    \label{fig:overview}
\end{figure*}

\section{Introduction}
Large Multimodal Models (LMMs) have achieved remarkable success on a broad spectrum of question-answering tasks in naturalistic settings, showcasing strong perceptual and reasoning abilities as well as extensive embedded knowledge~\citep{li2025perception, liu2026visionreasoner, zhou2025group}. 
Nevertheless, the limited evaluation of LMMs in industrial scenarios has hindered the conversion of this potential into practically deployable systems and tools~\citep{dou2026cl, yue2024mmmu}. 
Industrial environments pose unique challenges that are largely underexplored in current benchmarks, particularly due to safety-oriented task requirements and the relatively dense spatial distribution of targets.

Logistics warehousing is a representative industrial scenario in which continuous monitoring and timely safety risk recognition are crucial for accident prevention, regulatory compliance, and efficient warehouse operations. 
Previous studies have begun to develop multimodal benchmarks for industrial logistics, but their applicability to real-world assessment remains constrained by two inherent obstacles:
\textit{(i)} In terms of \textit{evaluation depth}, industrial logistics environments naturally raise concerns regarding commercial cost, privacy, safety, and liability. Consequently, IndustryEQA instead relies on physically simulated data produced within virtual environments. Collecting data from real-world industrial environments for systematic evaluation remains a significant gap.
\textit{(ii)} In terms of \textit{evaluation breadth}, practical deployment of LMMs involves diverse safety-critical dimensions, including object perception, spatial reasoning, path planning, and action recognition. However, ARMBench and iSafetyBench are limited to evaluating specific tasks under a single question format. Comprehensive multi-task evaluation of logistics safety-oriented perception and reasoning capabilities has yet to be established.
Therefore, the safety-oriented assessment of mainstream LMMs in real-world logistics scenarios remains an open question.

\begin{figure*}[!t]
    \centering
    \includegraphics[width=0.98\textwidth]{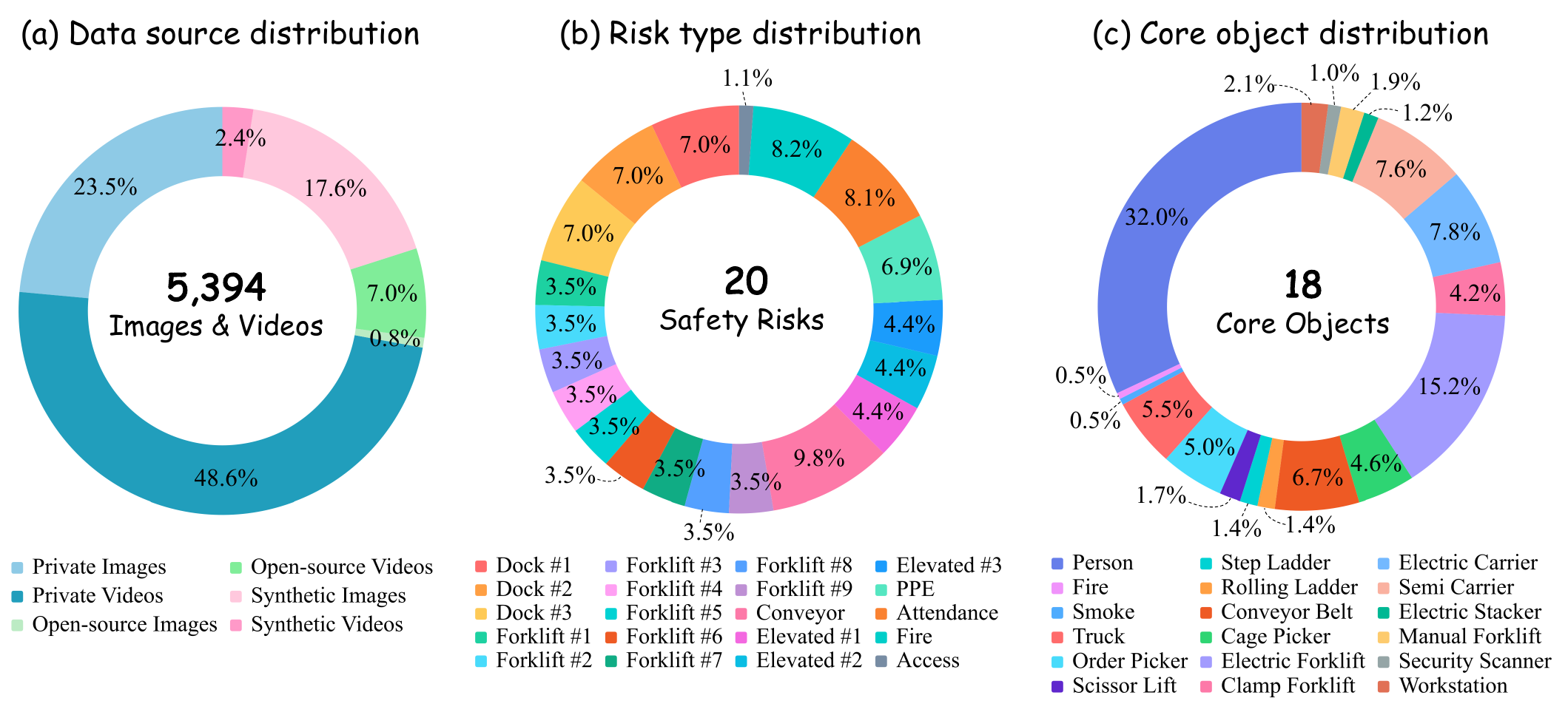}
    \caption{Core statistical distributions in \ourbench, comprehensively covering 5,394 visual samples, 20 safety risks, and 18 core objects.}
    \label{fig:statistics}
\end{figure*}

To systematically evaluate LMMs for logistics safety inspection, we introduce \textbf{\ourbench}, an industrial benchmark consisting of 2,476 images, 2,918 videos, 1,0274 VQAs, 20 risk factors, and 18 logistics objects.
As illustrated in Fig.~\ref{fig:overview}, we adopt a progressive curriculum to broadly assess the deployment value of LMMs in real-world industrial scenarios. 
For \textit{industrial element perception}, 3.5 million surveillance clips collected from the global intelligent logistics company Cainiao capture the dynamics and heterogeneous operational scenarios within logistics parks.
For \textit{warehouse knowledge understanding}, recruited logistics experts collectively devote 52 person-days to annotating essential factors, including spatial locations, object attributes, hazardous actions, etc.
For \textit{potential risk reasoning}, we additionally incorporate open-ended problems and an LLM-as-a-Judge evaluation protocol to assess LMMs' reliability and fairness in sophisticated thinking.
Coupled evaluation tasks pose fundamentally new challenges to LMMs in satisfying the demands of industrial logistics deployment.

By conducting a holistic evaluation of 20 mainstream open-source and advanced proprietary LMMs, we summarize the \textbf{key findings} as follows:
\textit{(a)} As exemplified by Qwen3.5-Plus, open-source LMMs closely track proprietary performance and achieve parity with GPT-5.5 and Claude-Opus-4.7.
Nevertheless, LMMs still have a pronounced performance disparity relative to human experts, indicating substantial room for improvement.
\textit{(b)} Industrial scenarios pose unprecedented challenges to fine-grained visual perception. LMMs consistently perform below expectations on the fine-grained industrial element perception tasks. Open-source models attain average accuracies of 0.39 on FG-S and 0.36 on FG-C.
\textit{(c)} Tasks involving industrial element perception and warehouse knowledge understanding are predominantly driven by visual cues. In contrast, potential risk reasoning requires models proficient in sophisticated thinking to identify risk factors.
\textit{(d)} Mainstream LMMs exhibit a pervasive risk-averse bias when responding to risk-prediction queries. Proprietary commercial models tend to adopt conservative responses, whereas open-source models provide more permissive and candid reports. 
Only 25\% of LMMs pass our fairness-criteria audit.
Collectively, these findings emphasize the challenges intrinsic to the \ourbench benchmark and delineate directions for future research and model improvement.

The curated benchmark bridges the last-mile from well-trained, simulation-validated models to practical application, providing actionable guidance for real‑world readiness.
For LMMs developed for large-scale industrial deployment, achieving strong performance on \ourbench is necessary to demonstrate holistic security expertise and expert-level perception and reasoning in operational environments.  
We hope the unique visual perception challenges, risk reasoning characteristics, and response bias tendencies introduced by \ourbench will provide useful insights for future work.


\begin{figure*}[!ht]
    \centering
    \includegraphics[width=0.98\textwidth]{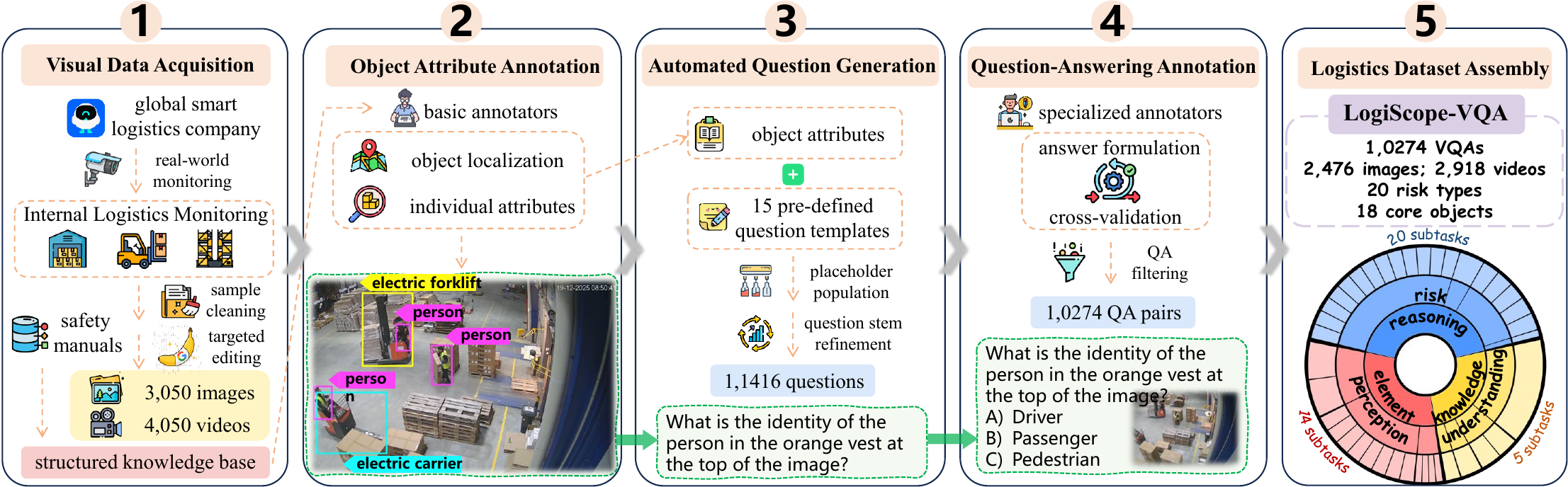}
    \caption{The curation pipeline of \ourbench, rigorously presenting the operational protocol alongside example outputs.}
    \vspace{-1em}
    \label{fig:curation_pipeline}
\end{figure*}

\section{\ourbench}
\label{Benchmark}

\subsection{Overview of Logistics Benchmark}
We introduce the \ourbench benchmark, a novel benchmark meticulously curated to evaluate the perception and reasoning capabilities of foundation LMMs for logistics safety in real-world industrial scenarios.
The visual corpus is derived from the warehouse surveillance of the global intelligent logistics company Cainiao, comprising 3.5 million raw clips that, after rigorous cleaning, filtering and augmentation, are distilled into 5,394 high-quality visual samples.
Logistics questions are automatically generated from human-annotated object attributes and hand-crafted question templates, comprehensively covering single-choice, multiple-choice, and open-ended formats.
Corresponding answers are authored, annotated, and cross-validated by recruited logistics experts through a hierarchical workflow, resulting in 10,274 high-quality VQA instances.
Collectively, the benchmark’s thorough data release and holistic experimental evaluation are underpinned by one year of surveillance across industrial warehouses worldwide, 52 person-days of annotation by logistics experts, and 324 million tokens consumed during LMM API inference.

Grounded in diverse visual content and granular attribute annotations, our benchmark encompasses single-choice, multiple-choice, and open-ended VQA problems.
As illustrated in Fig.~\ref{fig:statistics}, we achieve comprehensive coverage of key elements in industrial logistics scenarios, spanning 18 core objects and 20 risk types.
We aim to directly challenge LMMs to perceive logistics objects and integrate vertical knowledge to uncover latent industrial hazards.
Eventually, \ourbench is designed to rigorously evaluate the essential skills of LMMs, spanning industrial element perception, warehouse knowledge understanding, and potential risk reasoning.

Our benchmark proposes four key challenges for multimodal foundation models. Across diverse tasks, it requires LMMs to (i) perceive dense object distributions in industrial environments, (ii) acquire and operationalize logistics-specific domain knowledge, (iii) integrate multimodal information to perform reasoning, and (iv) anticipate and identify potential safety risks via sophisticated thinking.
We aim for the curated benchmark to serve as a comprehensive LMM testbed for identifying logistics safety risks in industrial scenarios, thereby precisely evaluating the real-world deployment potential of mainstream foundation models.


\subsection{Distinctive Positioning of \ourbench}
Distinct from existing benchmarks, we require LMMs to perform visual question-answering grounded in industrial contexts, enabling an intuitive assessment of practical operational value. 
The most closely related prior efforts are IndustryEQA~\citep{li2026industryeqa} and iSafetyBench~\citep{abdullah2025isafetybench}. 
However, IndustryEQA derives industrial data from simulated virtual environments, while iSafetyBench focuses solely on hazardous action recognition, constraining the breadth of visual elements and task coverage.
In contrast, \ourbench primarily sources data from real-world logistics warehouses, a nearly tenfold size over the previous dataset, and conducts a comprehensive multi-dimensional evaluation across perception, understanding, and reasoning.

\subsection{Dataset Curation Pipeline}
The construction of \ourbench followed a hierarchical curation pipeline that focused on data coverage and annotation quality, as illustrated in Fig.~\ref{fig:curation_pipeline}.
In \textbf{step 1}, we aggregated one year of surveillance video data from Cainiao’s global warehouse parks, amounting to 3.5 million raw clips. This massive corpus was then cleaned through video segmentation, camera-zone grouping, object detection pre-labeling and deduplication, yielding more than 7K high-quality visual samples as the initial seed data. 
Targeted editing is performed on approximately 900 samples with Nano Banana 2~\citep{gemini} to rebalance the distribution of rare risk types.
(The detailed image editing prompt is presented in  Appendix~\ref{image_editing}.) These over 7K seed samples were subsequently filtered in step 4 based on VQA pair quality, retaining the final 5,394 visual samples.
Concurrently, we identified core objects of interest within logistics safety and transformed their associated unstructured safety manuals into a structured schema.
In \textbf{step 2}, basic annotators spent 12 person-days annotating 18 core objects and their basic attributes. Specifically, they first drew bounding boxes to identify designated objects (\textit{e.g.}, persons, electric forklifts) on the keyframes, and then labeled the corresponding attributes (\textit{e.g.}, frame location, gender). To ensure accuracy, a third-party audit was conducted, with substandard annotations returned for iterative correction.
In \textbf{step 3}, the labeled object attributes were populated into the placeholders of the pre-defined 15 question templates to batch-generate initial VQA pairs, followed by a refinement operation. Such operation not only resolves referential ambiguity by adding specific modifiers and descriptive constraints to object names, but also improves sentence fluency by ensuring grammatical correctness and natural phrasing.
In \textbf{step 4}, four specialized annotators leveraged their domain expertise to annotate ground-truth answers and simultaneously filter out low-quality VQA pairs. This three-round cross-validation process, which took 40 person-days in total, comprised: 
(i) initial answer annotation and filtering of ill-formed pairs (\textit{i.e.}, those with ambiguous referents, unnatural phrasing, or missing options), which retained 5,394 visual samples from the initial 7K seed samples; 
(ii) independent re-annotation by a senior expert to accept concordant cases and adjudicate discordant cases with reference to the original labels to determine preferred answers;
and (iii) a stratified review that prioritized verifying cases with prior inter-round inconsistency. Consequently, approximately 90\% of the data was retained as high-quality VQA pairs.
In \textbf{step 5}, to ensure dataset diversity, we balanced the distribution of core objects, risk types, and hallucination rates by removing over-represented categories. Ultimately, the entire pipeline produced \ourbench, a comprehensive dataset comprising 10,274 high-quality VQA pairs. (Thorough details of the data processing pipeline are provided in Appendix~\ref{app:curation_details})

\begin{table*}[!ht]
    \caption{
    Main results on \ourbench across 10 evaluation tasks. 
    \textit{Abbr.}, 
    FP-S: Fine-grained Perception of Single-instance,
    FP-C: Fine-grained Perception of Cross-instance, 
    CP: Coarse-grained Perception; 
    WC: Warehouse Commonsense, 
    SR: Spatial Relation,
    OR: Operator Role; 
    PAC: Perimeter Access Control, 
    FM: Fire Monitoring, 
    PSD: Personnel Safety Duty,
    EOC: Equipment Operation Compliance.
    The optimal proprietary and open-source models are highlighted in \colorbox{backpurple}{purple} and \colorbox{backyellow}{yellow}, respectively.}
    \vspace{-1em}
    \label{tab:main_res}
    \begin{center}
    \setlength{\tabcolsep}{0.5em}
    \resizebox{\textwidth}{!}{
    \begin{tabular}{l | c | c c c | c c c | c c c c}
            \toprule
            \multirow{2}{*}{Model} & \multirow{2}{*}{Overall} & \multicolumn{3}{c|}{\ghead{Industrial Element\\Perception}} & \multicolumn{3}{c|}{\ghead{Warehouse Knowledge\\Understanding}} & \multicolumn{4}{c}{\ghead{Potential Risk\\Reasoning}} \\
            \cmidrule(lr){3-5} \cmidrule(lr){6-8} \cmidrule(lr){9-12}
             & & FP-S & FP-C & CP & WC & SR & OR & PAC & FM & PSD & EOC \\
            \midrule
            \multicolumn{12}{c}{\textit{Heuristics Baselines}} \\
            \midrule
            \icon{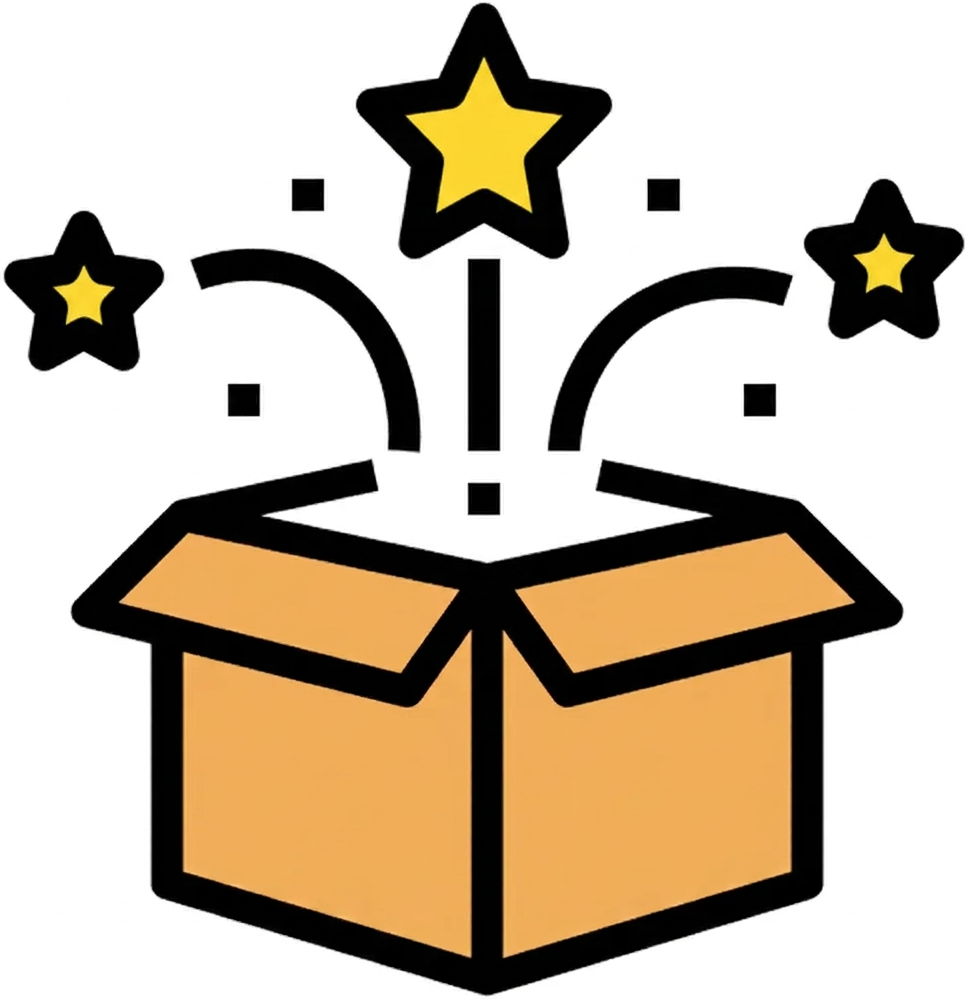} Random Choice & 0.33 & 0.24 & 0.22 & 0.32 & 0.31 & 0.15 & 0.07 & 0.37 & 0.46 & 0.46 & 0.47 \\
            \icon{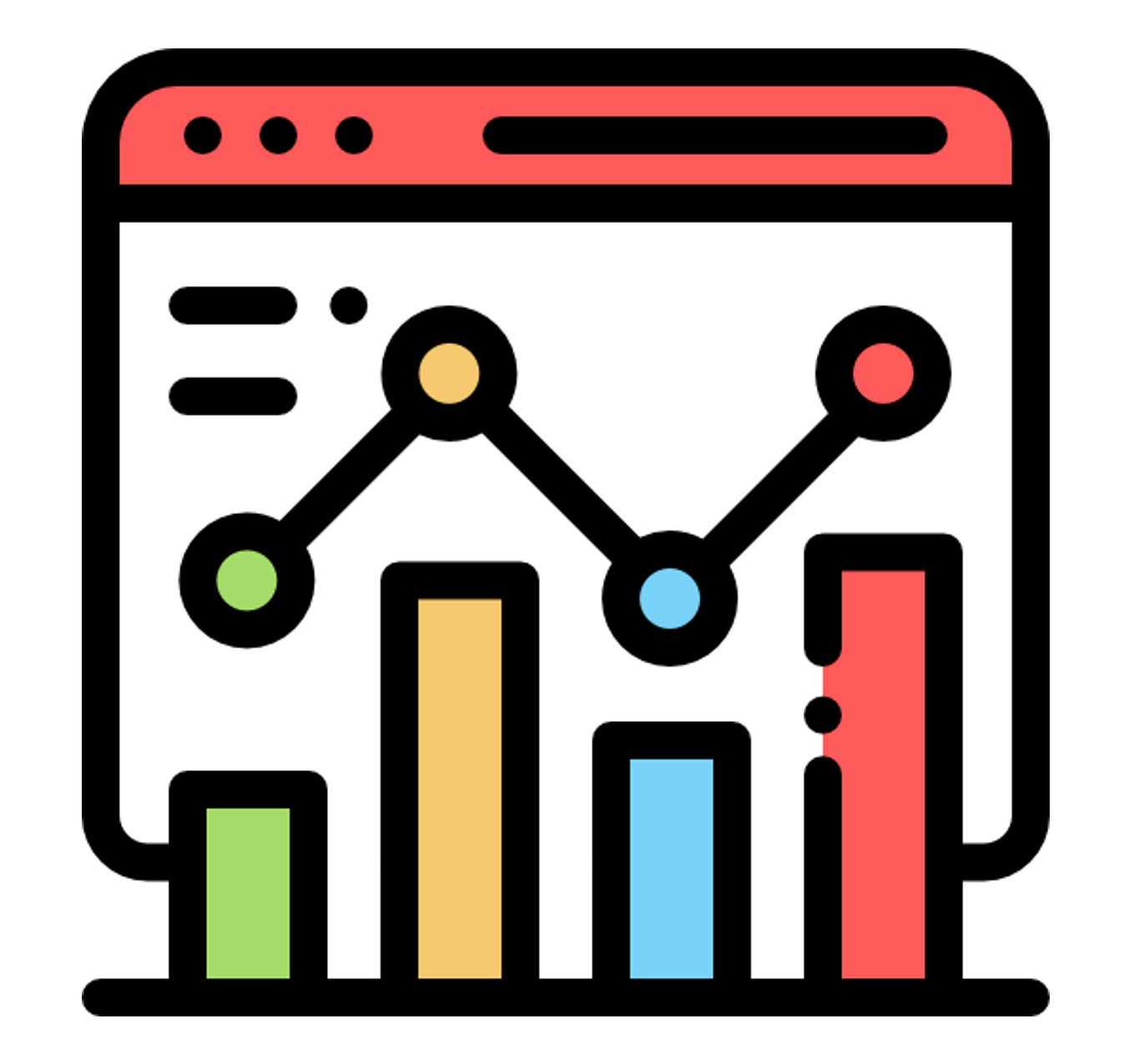} Frequent Guess & 0.47 & 0.32 & 0.27 & 0.63 & 0.39 & 0.24 & 0.40 & 0.74 & 0.49 & 0.64 & 0.80 \\
            \midrule
            \multicolumn{12}{c}{\textit{Proprietary Large Multimodal Models}} \\
            \midrule
            \icon{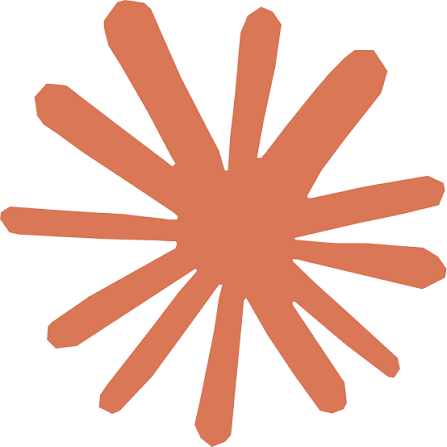} Claude-Sonnet-4.6 & 0.64 & 0.63 & 0.64 & 0.86 & 0.90 & 0.58 & 0.20 & 0.49 & 0.76 & 0.58 & 0.62 \\
            \icon{claude.png} Claude-Opus-4.7 & 0.70 & 0.64 & 0.61 & 0.88 & 0.89 & 0.60 & 0.30 & 0.74 & 0.82 & 0.65 & 0.79 \\
            \icon{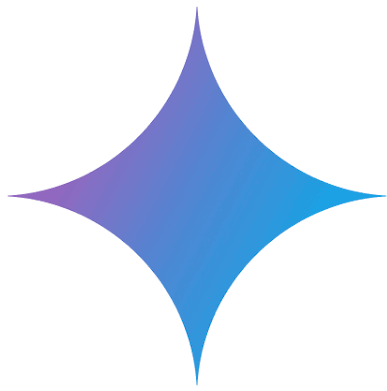} Gemini-3.1-Pro & 0.68 & \colorbox{backpurple}{0.72} & \colorbox{backpurple}{0.70} & \colorbox{backpurple}{0.91} & 0.92 & \colorbox{backpurple}{0.66} & 0.30 & 0.48 & 0.82 & 0.58 & 0.61 \\
            \icon{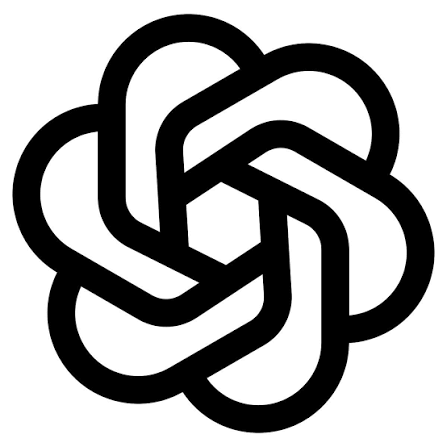} GPT-5.4 & 0.69 & 0.59 & 0.58 & 0.90 & 0.93 & 0.59 & 0.25 & 0.58 & \colorbox{backpurple}{0.84} & \colorbox{backpurple}{0.69} & 0.78 \\
            \icon{gpt.png} GPT-5.5 & \colorbox{backpurple}{0.71} & 0.62 & 0.68 & 0.78 & \colorbox{backpurple}{0.95} & 0.59 & 0.20 & \colorbox{backpurple}{0.79} & 0.82 & 0.65 & \colorbox{backpurple}{0.81} \\
            \icon{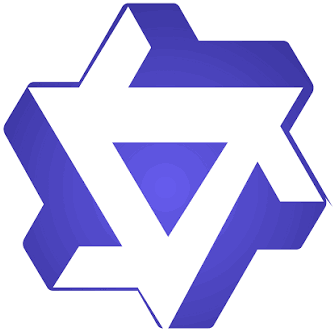} Qwen3.7-Plus & 0.66 & 0.67 & 0.64 & 0.89 & \colorbox{backpurple}{0.95} & 0.56 & \colorbox{backpurple}{0.40} & 0.48 & 0.83 & 0.53 & 0.69 \\
            \midrule
            \multicolumn{12}{c}{\textit{Open-source Large Multimodal Models}} \\
            \midrule
            \icon{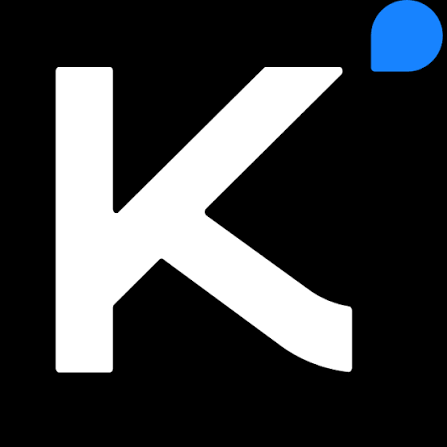} Kimi-K2-Thinking & 0.49 & 0.45 & 0.54 & 0.72 & \colorbox{backyellow}{0.96} & 0.18 & 0.05 & 0.15 & 0.30 & 0.29 & 0.14 \\
            \icon{kimi.png} Kimi-K2.6 & 0.62 & 0.61 & 0.60 & 0.89 & 0.93 & 0.45 & 0.20 & 0.57 & \colorbox{backyellow}{0.85} & 0.59 & 0.66 \\
            \icon{qwen.png} Qwen3-VL-8B & 0.50 & 0.48 & 0.44 & 0.89 & 0.87 & 0.09 & 0.05 & 0.42 & 0.73 & 0.57 & 0.65 \\
            \icon{qwen.png} Qwen3-VL-32B & 0.55 & 0.55 & 0.54 & 0.83 & 0.92 & 0.27 & 0.15 & 0.46 & 0.74 & 0.60 & 0.59 \\
            \icon{qwen.png} Qwen3-VL-235B & 0.64 & 0.64 & 0.59 & 0.89 & 0.91 & 0.55 & 0.25 & 0.43 & 0.73 & 0.60 & 0.67 \\
            \icon{qwen.png} Qwen3-VL-Plus & 0.34 & 0.53 & 0.45 & 0.76 & 0.91 & 0.15 & 0.05 & 0.20 & 0.57 & 0.30 & 0.13 \\
            \icon{qwen.png} Qwen3.5-Plus & \colorbox{backyellow}{0.70} & \colorbox{backyellow}{0.66} & \colorbox{backyellow}{0.65} & \colorbox{backyellow}{0.90} & 0.93 & \colorbox{backyellow}{0.61} & \colorbox{backyellow}{0.45} & 0.48 & 0.84 & 0.59 & 0.77 \\
            \icon{InternVL.jpeg} InternVL3.5 & 0.33 & 0.48 & 0.39 & 0.88 & 0.69 & 0.04 & 0.05 & 0.44 & 0.83 & 0.42 & 0.17 \\
            \icon{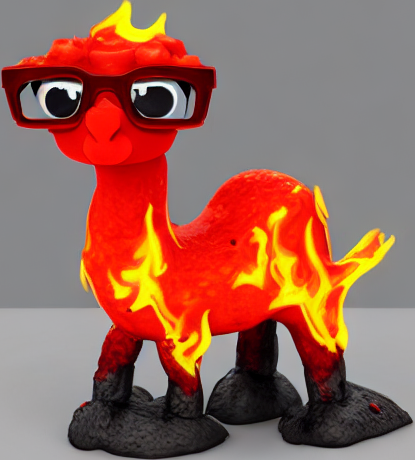} LLaVA-v1.6-7B & 0.50 & 0.34 & 0.30 & 0.48 & 0.60 & 0.21 & 0.05 & \colorbox{backyellow}{0.85} & 0.62 & 0.69 & \colorbox{backyellow}{0.82} \\
            \icon{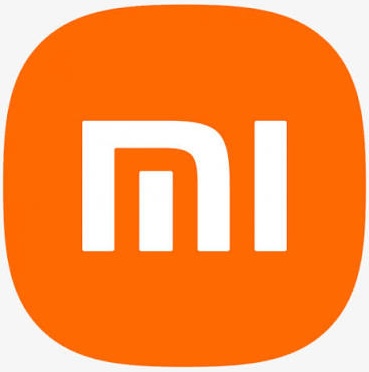} MiMo-VL-7B-RL & 0.44 & 0.46 & 0.39 & 0.65 & 0.32 & 0.05 & 0.10 & 0.44 & 0.74 & \colorbox{backyellow}{0.70} & 0.60 \\
            \icon{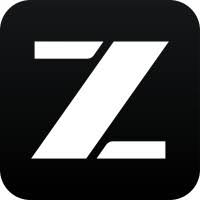} GLM-4.7 & 0.48 & 0.40 & 0.41 & 0.70 & 0.95 & 0.20 & 0.15 & 0.06 & 0.40 & 0.29 & 0.16 \\
            \icon{zhipu.jpeg} GLM-5.2 & 0.53 & 0.44 & 0.45 & 0.72 & 0.95 & 0.20 & 0.15 & 0.41 & 0.45 & 0.37 & 0.38 \\
            \icon{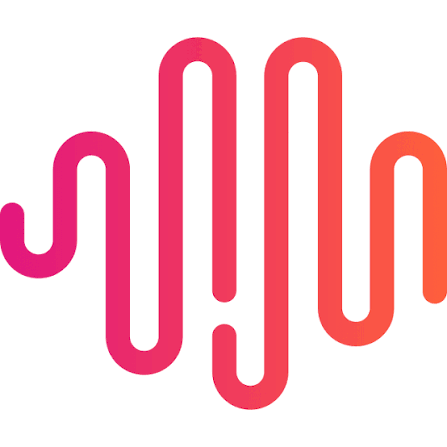} MiniMax-M2.1 & 0.43 & 0.47 & 0.49 & 0.54 & 0.93 & 0.19 & 0.15 & 0.50 & 0.43 & 0.40 & 0.60 \\
            \icon{minimax.png} MiniMax-M2.5 & 0.46 & 0.50 & 0.54 & 0.61 & 0.93 & 0.18 & 0.15 & 0.42 & 0.41 & 0.43 & 0.59 \\
            \midrule
            \multicolumn{12}{c}{\textit{Human Performance}} \\
            \midrule
            \icon{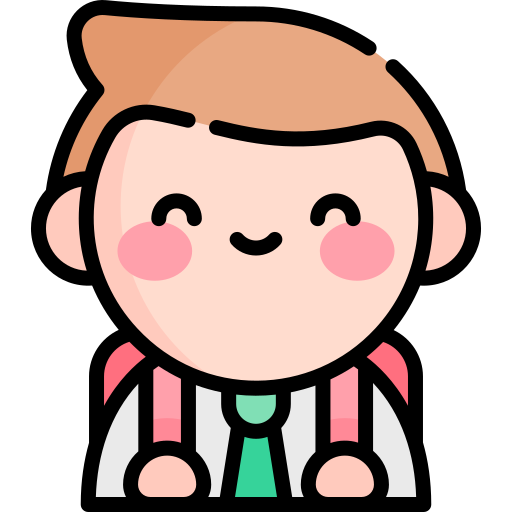}  Novice Student 
            & 0.73 & 0.80 & 0.81 & 0.85 & 0.76 & 0.58 & 0.75 & 0.75 & 0.80 & 0.70 & 0.67 \\ 
            \icon{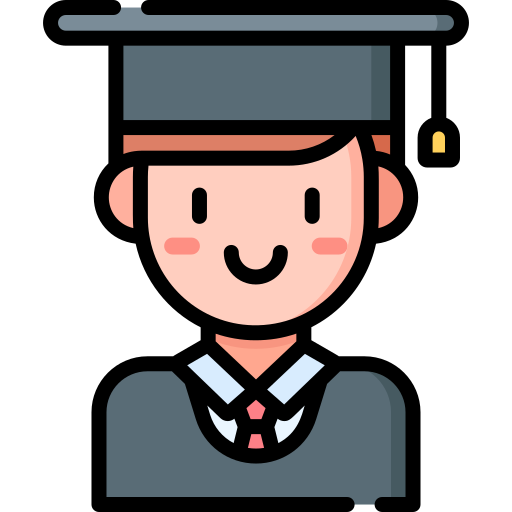}  Logistics Expert 
            & \colorbox{gray!25}{0.95} 
            & \colorbox{gray!25}{0.97} 
            & \colorbox{gray!25}{0.97} 
            & \colorbox{gray!25}{0.96} 
            & \colorbox{gray!25}{1.00} 
            & \colorbox{gray!25}{0.99} 
            & \colorbox{gray!25}{0.95} 
            & \colorbox{gray!25}{0.93} 
            & \colorbox{gray!25}{0.95} 
            & \colorbox{gray!25}{0.95}
            & \colorbox{gray!25}{0.91} \\
            \bottomrule
    \end{tabular}}
    \end{center}
    \vspace{-2em}
\end{table*}

\section{Experiment}
\label{Experiments}

\subsection{Evaluation Setups}
\label{Evaluation_Setups}

\paragraph{LMM Baselines}
We evaluate the difficulty of \ourbench using diverse state-of-the-art LMMs as baselines and establish robust reference points for future research. For proprietary LMMs, six leading commercial models are selected: Claude-Sonnet-4.6~\citep{sonnet4.6}, Claude-Opus-4.7~\citep{opus4.7}, Gemini-3.1-Pro~\citep{gemini}, GPT-5.4, GPT-5.5~\citep{gpt5}, and Qwen3.7-Plus~\cite{qwen3.7}. These models represent the current upper bound of closed-source multimodal capabilities and serve as strong performance ceilings for \ourbench. For open-source LMMs, models spanning multiple architectural families and parameter scales are included: MiMo-VL-7B-RL~\citep{mimo-vl}, GLM-5.2~\citep{glm5.2}, GLM-4.7~\citep{glm4.7}, Kimi-K2-Thinking~\citep{kimi-k2-thinking}, Kimi-K2.6~\citep{kimi-k2.6}, InternVL-3.5~\citep{internvl3.5}, LLaVA-v1.6-7B~\citep{llava-v1.6}, Qwen3.5-Plus~\citep{qwen3.5}, and multiple Qwen3-VL variants (8B, 32B, 235B-A22B, Plus)~\citep{qwen3}.
Moreover, we incorporate the Random Choice and Frequent Guess baselines to provide chance-level and prior-biased reference points. While computed in the standard manner for multiple-choice questions, the Random Choice accuracy is treated as zero for open-ended questions, given the extremely low probability that a solution randomly sampled from the vast candidate space would be consistent with the ground-truth answer.
Meanwhile, the Frequent Guess baseline uses the most frequent canonical answer within each task category as the prediction for all open-ended questions.

\paragraph{Human Performance}
To assess holistic human performance, we recruited ten college students with no expertise in industrial logistics to represent the novice level, and one senior logistics specialist with over four years of on-site warehouse safety experience to represent the expert level.
Human participants completed all questions independently based solely on professional knowledge, without external references or model outputs. This setup ensures that the reported performance reflects genuine domain expertise rather than test-taking strategies or external aids, serving as a reliable upper bound for model evaluation.
(The detailed employment terms for human evaluators are provided in the Appendix~\ref{app:human_eval_protocol}.)

\paragraph{Decoding Settings}
For all problem instances, we use a unified prompt template that explicitly instructs the models to produce answers in a standardized format. For single-choice and multiple-choice questions, correctness is determined via direct keyword matching. For open-ended questions, we develop an LLM-as-a-Judge workflow to extract the key conclusion from each response for answer matching. Eventually, we adopt micro-averaged accuracy as the evaluation metric.
All evaluated models are accessed via their official APIs or publicly available checkpoints. For each task, we repeat each test three times and report the average performance to ensure reliability and reproducibility. The sampling temperature for each model is set to its recommended or default value. 
(Detailed prompts and decoding setups are in Appendix~\ref{app:decoding_setups}.)


\subsection{Overall Results}
\subsubsection{Main Performance Analysis}
As reported in Tab.~\ref{tab:main_res}, proprietary models maintain an advantage, yet the gap with open-source models has essentially closed at the top, where the best open-source model now ties the strongest proprietary systems. These leading models have edged past the untrained human (novice student) while remaining far below the logistics expert, confirming that our benchmark measures learnable professional expertise rather than common intuition. 
The key finding, however, is that fine-grained perception in logistics safety scenarios constitutes a shared, industry-wide bottleneck. All top models cluster within a narrow, low band on FG-S and FG-C, falling short of even the novice human and trailing the expert by a wide margin. This is partly because surveillance images in these settings are typically wide-angle, low-resolution, and densely cluttered with numerous targets and backgrounds, posing a far greater challenge than the clean, high-quality images common in standard benchmarks. In stark contrast, on knowledge-intensive tasks such as AR and EOC, models already outperform untrained humans. This contrast reveals an interesting structural complementarity between human and machine capabilities. Humans naturally excel at spotting fine-grained visual details in logistics safety scenarios even without training, whereas models, despite their vast knowledge, often behave like a well-read scholar with poor eyesight.

\subsubsection{In-depth Thinking Analysis}
\begin{figure*}[htbp]
    \centering
    \begin{minipage}[c]{0.65\textwidth}
        \centering
        \captionof{table}{Cross-task performance comparison with and without thinking mode. Colored cells indicate significant changes.}
        \vspace{-0.5em}
        \label{tab:think_vs_nothink}
        \resizebox{\linewidth}{!}{%
        \begin{tabular}{l|cc|cc|cc}
            \toprule
            Model & 
            \multicolumn{2}{c|}{\ghead{Industrial Element\\Perception}} & 
            \multicolumn{2}{c|}{\ghead{Warehouse Knowledge\\Understanding}} & 
            \multicolumn{2}{c}{\ghead{Potential Risk\\Reasoning}} \\
            \cmidrule(lr){2-3} \cmidrule(lr){4-5} \cmidrule(lr){6-7}
            & non-think & think & non-think & think & non-think & think \\
            \midrule
            LLaVA-v1.6-7B & \obg{0.28} & \obg{0.30\hspace{0.25em}{\scriptsize\textcolor{red}{$\uparrow$7.1\%}}} & \rbg{0.27} & \rbg{0.31\hspace{0.25em}{\scriptsize\textcolor{red}{$\uparrow$14.8\%}}} & \obg{0.65} & \obg{0.71\hspace{0.25em}{\scriptsize\textcolor{red}{$\uparrow$9.2\%}}} \\
            Kimi-K2.6 & 0.66 & 0.66 & \bbg{0.71} & \bbg{0.57\hspace{0.25em}{\scriptsize\textcolor{red}{$\downarrow$19.7\%}}} & \obg{0.68} & \obg{0.74\hspace{0.25em}{\scriptsize\textcolor{red}{$\uparrow$8.8\%}}} \\
            MiniMax-M2.1 & \bbg{0.14} & \bbg{0.10\hspace{0.25em}{\scriptsize\textcolor{red}{$\downarrow$28.6\%}}} & 0.38 & 0.38 & 0.62 & 0.63 \\
            GLM-5.2 & \bbg{0.15} & \bbg{0.12\hspace{0.25em}{\scriptsize\textcolor{red}{$\downarrow$20.0\%}}} & 0.39 & 0.39 & \obg{0.50} & \obg{0.54\hspace{0.25em}{\scriptsize\textcolor{red}{$\uparrow$8.0\%}}} \\
            Qwen3-VL-235B & 0.63 & 0.64 & 0.66 & 0.69 & 0.70 & 0.71 \\
            Qwen3.5-Plus & \bbg{0.69} & \bbg{0.68\hspace{0.25em}{\scriptsize\textcolor{red}{$\downarrow$1.5\%}}} & 0.72 & 0.74 & \rbg{0.72} & \rbg{0.80\hspace{0.25em}{\scriptsize\textcolor{red}{$\uparrow$11.1\%}}} \\
            Qwen3.7-Plus & 0.69 & 0.69 & \bbg{0.71} & \bbg{0.68\hspace{0.25em}{\scriptsize\textcolor{red}{$\downarrow$4.2\%}}} & \rbg{0.69} & \rbg{0.77\hspace{0.25em}{\scriptsize\textcolor{red}{$\uparrow$11.6\%}}} \\
            \bottomrule
        \end{tabular}
        }
    \end{minipage}
    \hfill
    \begin{minipage}[c]{0.34\textwidth}
        \centering
        \includegraphics[width=\linewidth]{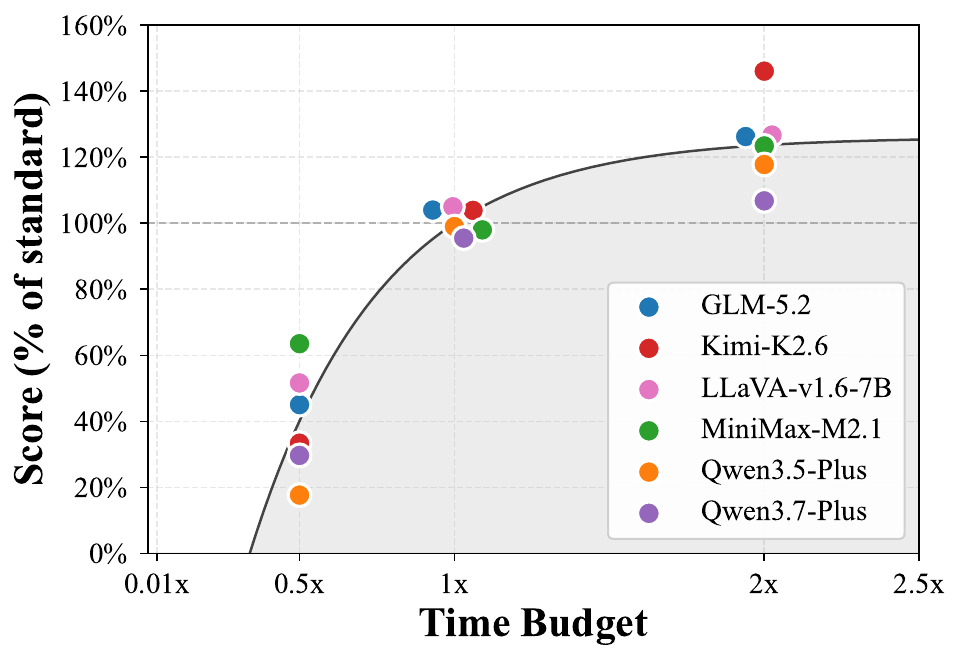}
        \vspace{-2em}
        \captionof{figure}{Impact of thinking time budget on reasoning-driven tasks.}
        \label{fig:think_budget}
    \end{minipage}
\end{figure*}

As shown in Table~\ref{tab:think_vs_nothink} (evaluated on a subset of \ourbench), red, orange and blue cells present notable increases, moderate increases and noticeable drops, respectively. We can see that enabling thinking mode yields a mixed effect rather than a uniform improvement. Across models, the overall ranking remains largely stable, but the performance shift is highly task-dependent, suggesting that thinking mode interacts differently with perception-heavy and reasoning-heavy tasks. \textit{Industrial Element Perception} is the least responsive dimension, while \textit{Warehouse Knowledge Understanding} shows modest gains, which may be constrained by the perceptual layer itself. However, \textit{Potential Risk Reasoning} benefits the most, especially for stronger models such as Qwen3.5-Plus. This is because the process of observing the scene, matching safety rules and drawing inferences is reasoning-dominant, making it particularly well aligned with explicit chain-of-thought enabled by thinking mode.


Building on this observation, we further probe whether the reasoning advantage persists as the thinking time budget increases for the same \textit{Potential Risk Reasoning} VQA samples.
Figure~\ref{fig:think_budget} indicates a clear saturation pattern across all models. At low budgets, performance degrades severely due to output truncation, while further gains become marginal beyond the standard budget.
Therefore, to balance performance and token consumption, thinking mode should be selectively enabled depending on the task's primary reliance on visual perception or reasoning. And the time budget should be set near the saturation point.

\subsubsection{Safety Risk Bias Analysis}
\begin{figure*}[ht]  
    \centering
    \includegraphics[width=1.0\textwidth]{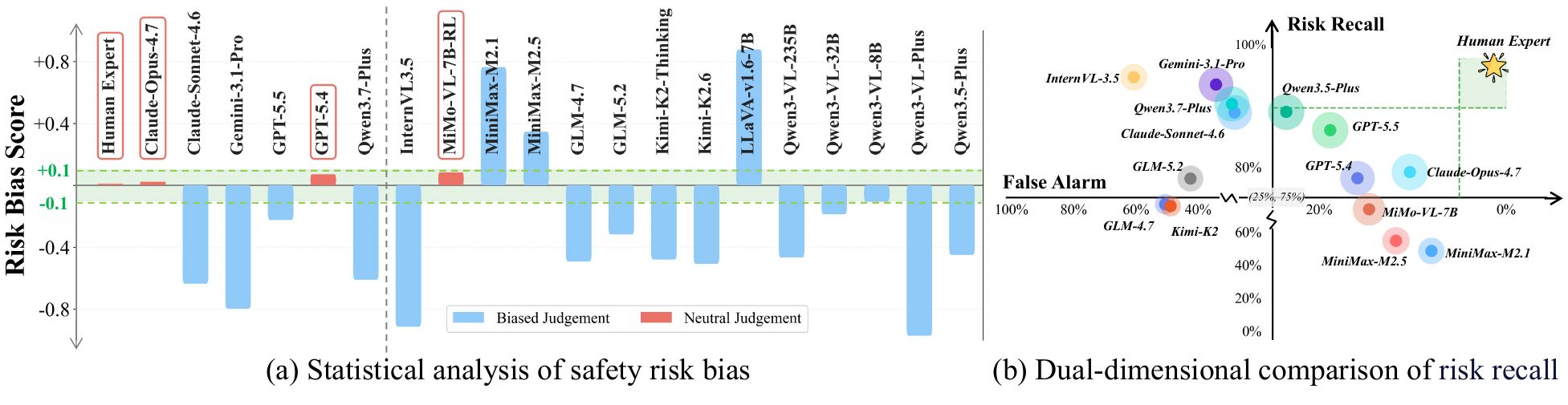}
    \caption{Empirical analysis of safety risk biases in mainstream LMMs.
    Panel (a) depicts holistic characterization of deviations in risk bias scores.
    Panel (b) provides detailed comparison across risk‑recall and false‑alarm dimensions.}
    \label{fig:recall_bias}
    \vspace{-1em}
\end{figure*}

We investigate whether models' safety risk judgments exhibit systematic bias or random noise with two experiments, as shown in Fig.~\ref{fig:recall_bias}. In panel (a), we quantify global directional skew via the indicator RBS, defined as $\mathrm{Risk\ Bias\ Score} = \tanh\!\left(\frac{\log((\mathrm{FNR}+\epsilon)/(\mathrm{FPR}+\epsilon))}{c}\right)$,  where FNR and FPR denote the false negative rate and false positive rate, and $\epsilon=10^{-8}$ and $c=2$. Positive, negative, and near-zero RBS indicate optimistic, conservative, and neutral judgments, respectively.
Our evaluation reveals two critical findings. \textbf{First, conservative bias dominates}, with most models exhibiting a negative RBS. This systematic over-reporting tendency likely reflects a safety-first alignment strategy that prioritizes high recall, penalizing missed detections more heavily than false alarms. 
\textbf{Second, proprietary and open-source models exhibit a marked performance divergence, suggesting a cognitive divide between genuine risk comprehension and superficial pattern matching}. While top proprietary models cluster near the unbiased neutral zone, open-source ones scatter across the entire score range. This pattern implies that elite proprietary models may leverage event-level risk reasoning to accurately distinguish benign events from hazards, whereas less capable models lack such depth. Consequently, the latter likely rely on defensive over-reporting to mitigate uncertainty, manifesting as systemic conservative bias.

Building on this global picture, we conduct a paired-image probe to examine how such bias behaves under explicit risk removal. Specifically, we evaluate models on 100 surveillance samples containing genuine hazards and a corresponding set of risk-free samples, captured by the same cameras at a different time period, to measure recall and false alarm rates, respectively. As depicted in panel (b), most models maintain a risk recall above 60\% but their false alarm rates vary substantially, confirming the prevalence of conservative bias. Moreover, The models are distributed across three quadrants, with those in quadrant 2 exhibiting excessive sensitivity and high false alarm rates, whereas those in quadrant 4 underperform by missing genuine threats. In contrast, models in quadrant 1 achieve the optimal trade-off between recall and false alarm rates, yet they still remain far from human expert performance. Collectively, our results underscore that safety risk bias is a pervasive systemic issue, which may be attributed to an interplay between internal cognitive limitations and external factors such as training data and alignment protocols. 




\section{Challenges and Future Directions}
Evaluations on \ourbench expose two key limitations of current models. Firstly, fine-grained perception in logistics safety scenarios remains a major bottleneck, where models particularly struggle with low-resolution, wide-angle, heavily occluded and densely cluttered surveillance footage. To address this, future efforts should aim to boost models' intrinsic visual capacity, achievable through strategies like optimizing pre-training for dense and low-resolution targets, or exploiting temporal information across video frames. Secondly, safety risk bias emerges as a significant yet overlooked issue,
characterized by a prevalent unfairness, with most models heavily inclined to over-report risks. This problem deserves more attention from model developers, as industrial application models require fairness as much as accuracy. For future work, a crucial direction is to prevent such bias at the source, for example through balanced data curation or bias-aware training objectives.

\section{Related Work}
\subsection{LMM Industrial Assessment}
Industrial logistics safety assessment provides a compelling testbed for studying the practical applicability of LMMs. Unlike natural-scene evaluation~\citep{liu2024mmbench, OpenEQA2023}, industrial environments~\citep{mitash2023armbench, abdullah2025isafetybench} impose the integration of sophisticated object recognition, precise spatial reasoning, and safety-critical hazard identification. 
However, this research area remains heavily underexplored, as industrial data source are often constrained by commercial confidentiality.
IndustryEQA~\citep{li2026industryeqa} is a pioneering study on logistics safety evaluation in warehouse scenarios. However, it contains only approximately 1.3 thousand physics-simulated samples generated in virtual environments.
To fully alleviate this challenge, we release \ourbench, which consists of 10,274 VQA instances annotated by logistics experts. 
We expect the constructed benchmark to provide a direct and practical testbed for assessing the potential of LMMs in industrial applications.

\subsection{Practical Application Benchmarks}
LMMs have emerged as a major frontier in AI research, offering a unified modeling framework that integrates visual perception, language modeling, and sophisticated reasoning~\citep{radford2021learning, alayrac2022flamingo, liu2023visual}.
Through vision-language pre-training, instruction tuning, and reinforcement learning for reasoning, cutting-edge research has progressively advanced model capabilities across a broad spectrum of vertical domains.  
To keep pace with these advances, recent benchmarks have introduced targeted evaluations for 
reasoning-centric mathematical problem solving~\citep{yue2024mmmu, wang2026livek12bench, lu2024mathvista}, 
knowledge-intensive medical diagnosis~\citep{chen2024gmai, zhou2025protclip, liu2025gemex, yin2026caduceus}, 
and visually-grounded remote sensing tasks~\citep{wang2025xlrsbench, danish2025geobenchvlm, wang2025omniearth}.
Logistics safety evaluation focuses on workflows, equipment states, human behaviors and safety regulations, framing an embodied risk-oriented QA task in real industrial settings with distinctive domain-specific challenges.
\ourbench is the first dedicated evaluation suite for logistics scenarios, filling a critical gap in existing research.

\section{Conclusion}
The proposed \ourbench represents a significant advance in assessing the capabilities of LMMs for logistics hazard identification in industrial scenarios. 
Concretely, we comprehensively evaluate the basic skills of industrial element perception, warehouse knowledge understanding, and potential risk reasoning.
Extensive empirical evaluations reveal that LMMs fall short of expectations in industrial visual perception, and effective hazard identification hinges on coherent sophisticated reasoning. 
Moreover, our investigation uncovers a pervasive issue of security-related bias in current LMMs. 
These discrepant findings offer valuable insights for foundation model construction and practical deployment.
In future work, we will further enrich the visual question-answering corpus and extend the benchmark for agentic industrial operations, continuously contributing robust resources to the multimodal evaluation research community.

\section*{Acknowledgments}
We thank Yuan Zhou and other experts for their professional guidance on logistics safety knowledge and data annotation, and Qian Xu and her colleagues for their support in data annotation. We also thank the Qwen team for an insightful technical exchange, which deepened our understanding of the scarcity of logistics surveillance data for training foundation models. We have since shared a portion of our surveillance data with the team, in the hope that it may serve as a useful resource for future model development.

\bibliographystyle{abbrvnat}
\bibliography{main}

\newpage
\appendix
\section{Large Language Model Usage}
The conceptual innovation and motivation underlying this study are conceived independently by the human authors, without cognitive input from large language models. 
The manuscript is originally written by the human authors, and large language models are engaged only at the final stage to assist in polishing key academic terminology.
During dataset construction, we primarily relied on proprietary real-world data, while approximately 20.5\% of the dataset is generated by the large language model (\textit{i.e.}, Nano Banana 2~\citep{gemini}) to systematically enrich the diversity of risk factors.
For experimental evaluation, we employ large language models exclusively through the official API endpoints provided by respective vendors. All human–AI interactions are conducted in full compliance with all applicable terms of service and licensing conditions. 

\section{Ethics Statement}
We are committed to responsible AI research and strict adherence to data privacy standards. For the private data in \ourbench, we have implemented rigorous anonymization protocols. Specifically, all privacy-sensitive content—including clearly visible human faces, specific warehouse names and commercial identifiers—has been blurred or masked. This process combines automated detection with manual verification, ensuring that no individual or proprietary entity can be identified. In contrast, since the synthetic data contains no real-world identities and the open-source data is already publicly available, both are released in their original format without additional anonymization. We confirm that the construction and release of \ourbench comply with relevant ethical guidelines and data protection regulations.

\section{Data Usage and Licensing}
\ourbench is released under the \textbf{Creative Commons Attribution-NonCommercial-ShareAlike 4.0 International License (CC BY-NC-SA 4.0)}. It is strictly intended for academic and non-commercial research purposes only. Users are prohibited from using this data for any commercial activities, including but not limited to training commercial models, selling derived products, or integrating it into profit-generating services. Furthermore, any derivative works based on this dataset must be distributed under the same license. By accessing this dataset, users agree to adhere to these terms.

\begin{figure*}[htbp]  
    \centering
    \includegraphics[width=0.95\textwidth]{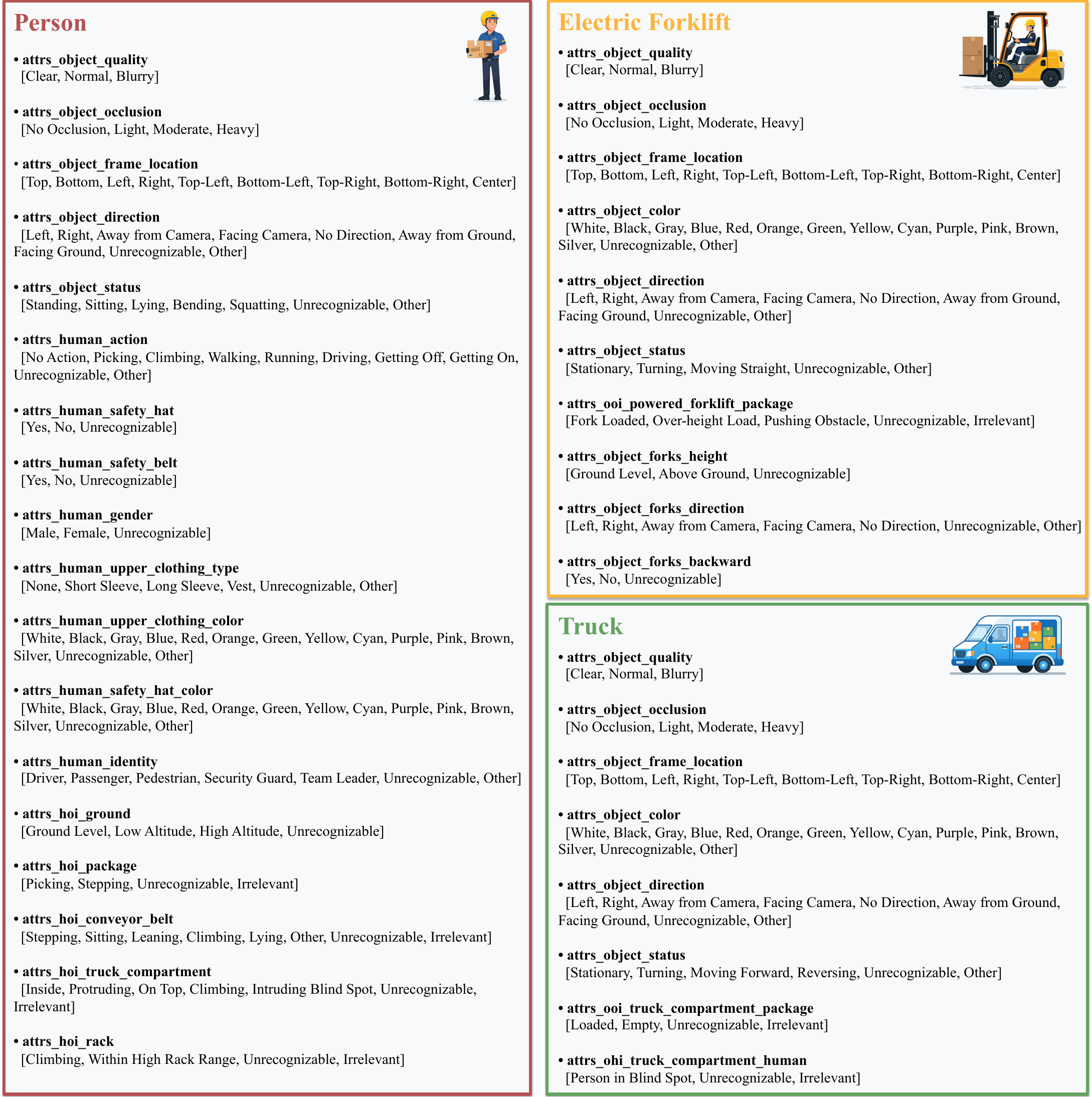}
    \caption{JSON schema examples for \textit{person}, \textit{truck} and \textit{electric forklift}.}
    \label{fig:schema}
\end{figure*}
\begin{figure*}[htbp]  
    \centering
    \includegraphics[width=0.95\textwidth]{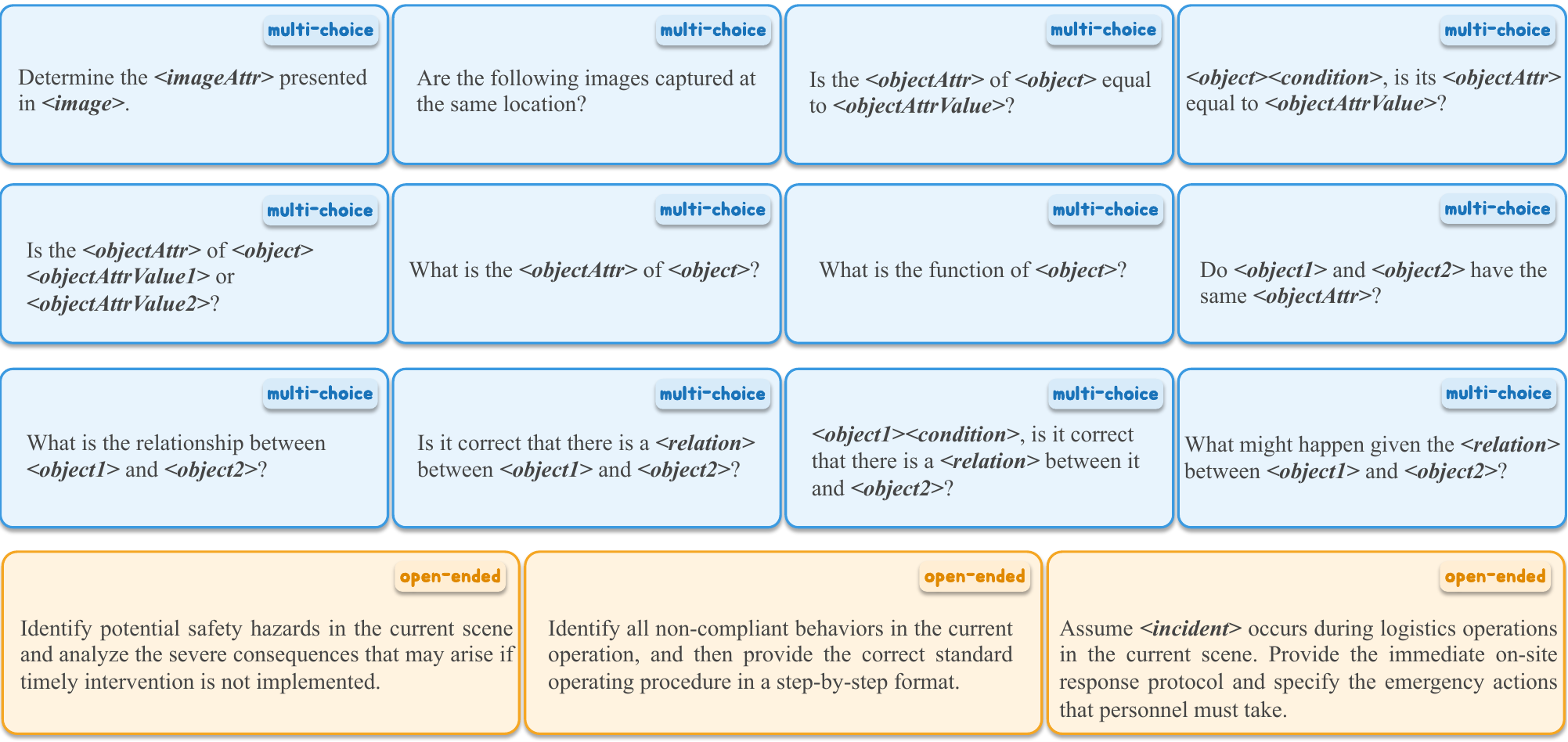}
    \caption{Question templates for generating question stems.}
    \label{fig:question_templates}
\end{figure*}

\begin{figure*}[htbp]  
    \centering
    \includegraphics[width=1.0\textwidth]{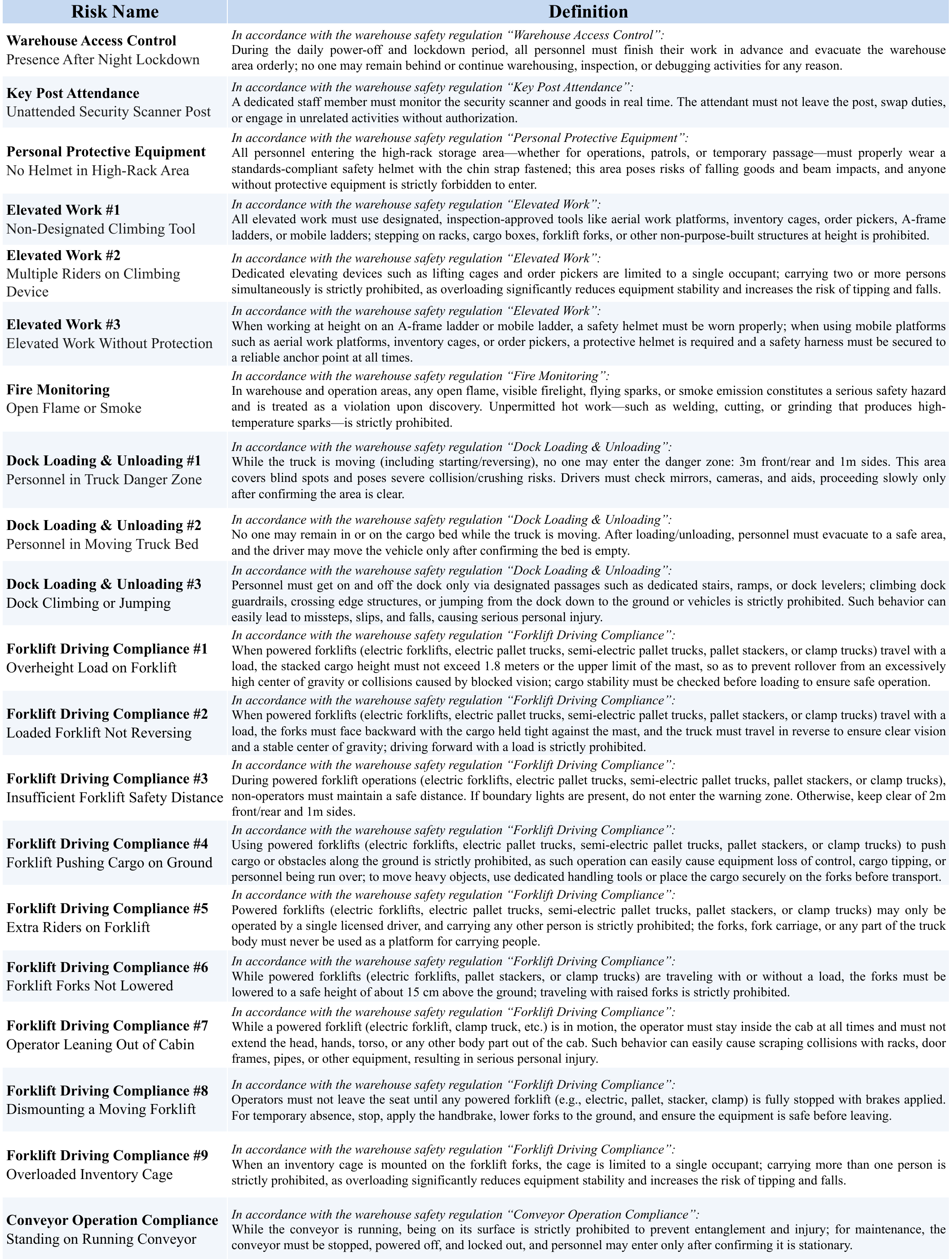}
    \caption{Definitions of 20 risk in potential risk reasoning.}
    \label{fig:risk_definition}
\end{figure*}

\begin{figure*}[htbp]  
    \centering
    \includegraphics[width=1.0\textwidth]{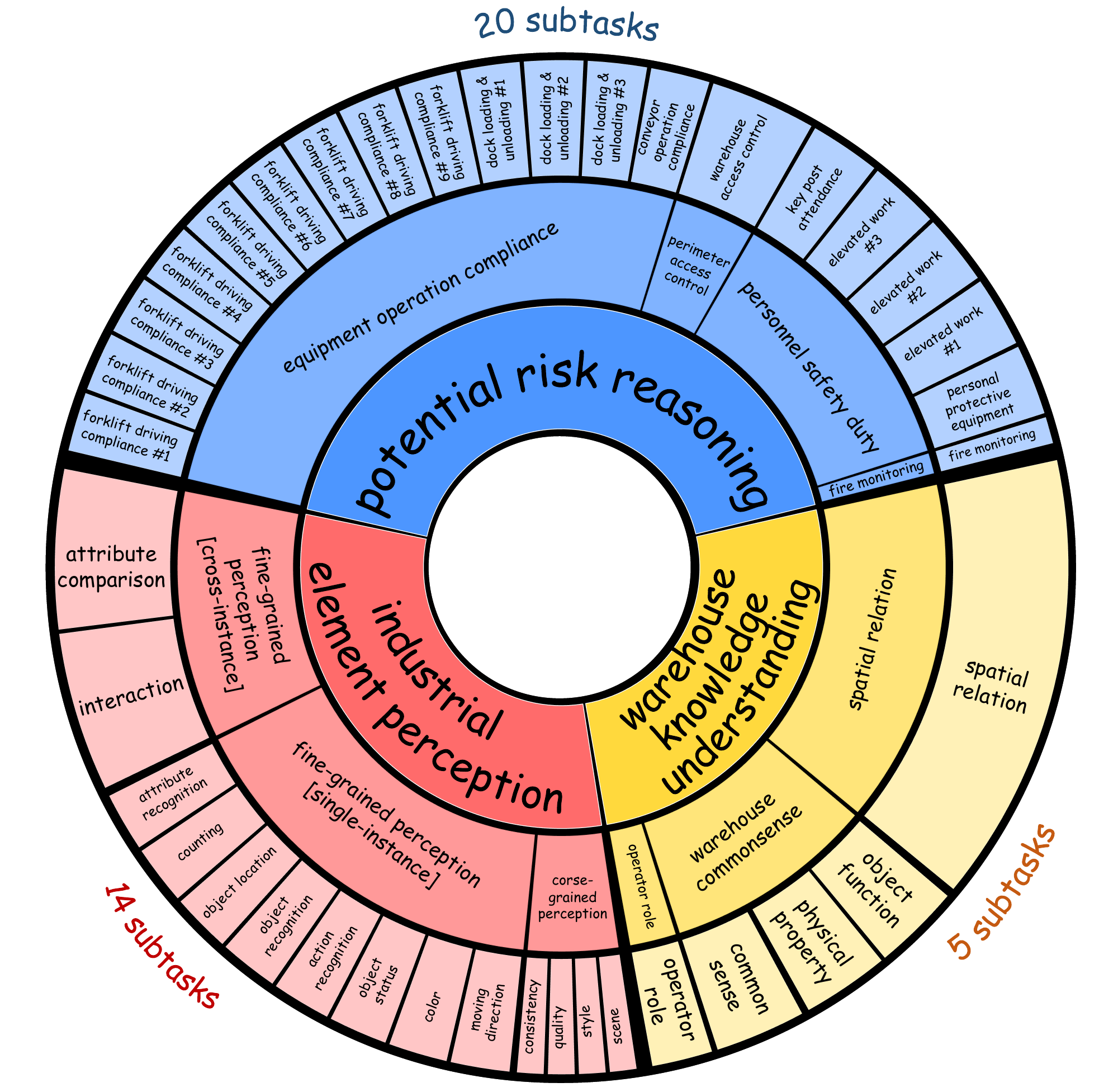}
    \caption{Full task taxonomy of \ourbench.}
    \label{fig:task_taxonomy}
\end{figure*}

\begin{figure*}[htbp]  
    \centering
    \includegraphics[width=1.0\textwidth]{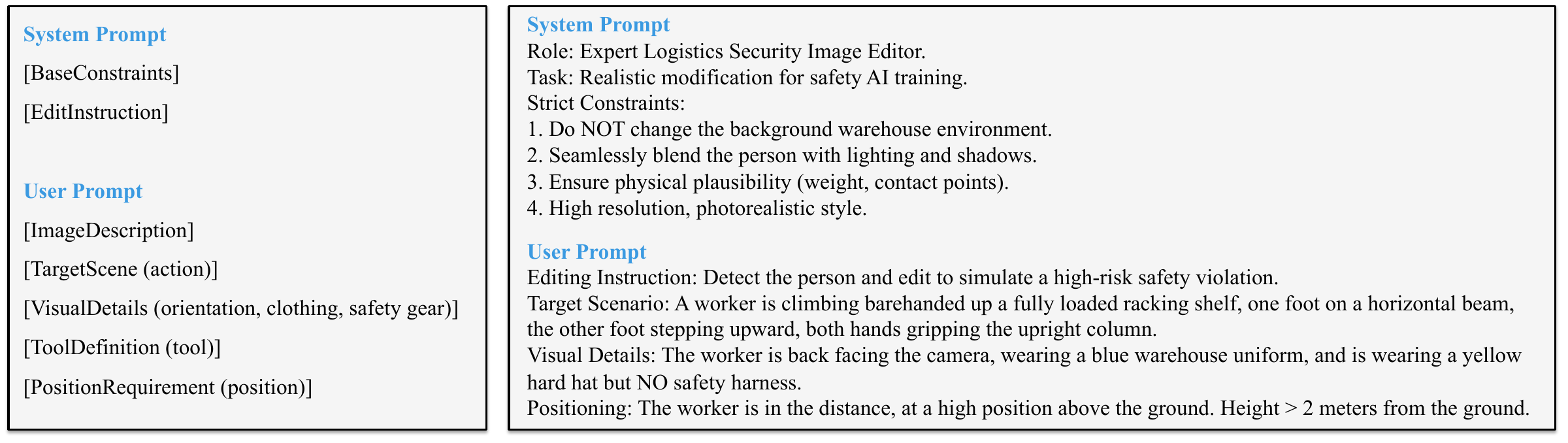}
    \caption{Prompt template and example for targeted image editing.}
    \label{fig:gen_prompt_template}
\end{figure*}

\section{Curation Details}
\label{app:curation_details}
\subsection{Hierarchical Curation Pipeline}
\paragraph{Visual Data Acquisition} 
The raw surveillance corpus comprises 3.5 million video clips collected from Cainiao’s global warehouse parks between April 15, 2025 and April 10, 2026. The data spans diverse operational scenarios, including high rack areas, loading docks, storage zones, light-duty picking shelves, office areas, etc. Notably, we declare that \textbf{no surveillance data from the United States was included in this collection}. To transform this massive raw corpus into a usable foundation, we implemented a rigorous cleaning process. Continuous videos were first segmented into clips or keyframes via fixed-interval sampling or risk-triggered alarm retrieval. These segments were then grouped by camera zones to align with specific operational areas. Subsequently, a pre-trained object detection model identified core objects, allowing us to filter out irrelevant background frames and balance object category distributions. The process concluded with deduplication, yielding over 7,000 high-quality, unique visual samples that serve as the foundational seed data for benchmark construction.
With this visual foundation in place, a structured knowledge base from unstructured safety manuals became essential. Specifically, logistics specifications are parsed into hierarchical JSON schemas based on semantic tuples (\textbf{\texttt{Object}},\textbf{\texttt{Attribute}},\textbf{\texttt{Value}}). In this formulation, \textbf{\texttt{Object}} denotes one of the 18 core entities critical to warehouse safety monitoring, such as person, electric carrier and security scanner. \textbf{\texttt{Attribute}} captures either the intrinsic state of an object (\textit{e.g.}, posture, PPE compliance) or its interaction relationship with other entities (\textit{e.g.}, human-truck proximity). \textbf{\texttt{Value}} consists of predefined enumerations derived from domain regulations, strictly constraining the attribute space. For instance, a valid tuple might be \texttt{(person, attrs\_hoi\_rack, climbing)}, which explicitly encodes the specific interaction of a person climbing a rack. Some examples of the JSON schema are shown in Fig.~\ref{fig:schema}.

\paragraph{Object Attribute Annotation} 
In Object Attribute Annotation, visual grounding was established via lightweight, cost-effective manual labeling. Specifically, annotators followed a "box-then-attribute" protocol: they first drew bounding boxes to localize core objects, and then marked basic attributes (e.g., frame location, occlusion status). For video samples, this process was applied exclusively to the single keyframe exhibiting the most salient risk characteristics, thereby maximizing annotation efficiency. Furthermore, to enrich semantic details, the "person" object received extended tags such as gender and upper-garment color. Crucially, all labels were constrained to predefined values aligned with the logistics knowledge base. To ensure final precision, a third-party quality audit was conducted, involving iterative sampling and correction cycles to rectify any remaining inaccuracies.

\paragraph{Automated Question Generation} 
This step transformed object attribute annotations into natural language queries through two sequential operations. First, we leveraged a library of 15 predefined question templates (Fig.~\ref{fig:question_templates}) to generate raw question stems. Specifically, twelve multiple-choice templates were designed to assess cognitive understanding across dimensions such as holistic scene interpretation, single-object recognition, and multi-object interaction. The remaining three open-ended templates evaluated capabilities in safety risk perception, consequence prediction, hazard rectification, and emergency response planning. Based on these templates, we performed deterministic population by mapping extracted core objects and attribute values into the corresponding slots. Second, we applied rule-driven linguistic refinement to enhance naturalness and precision. This refinement operation served two purposes. It synthesized multi-attribute constraints into natural pre-nominal modifiers to disambiguate and precisely localize target objects (\textit{e.g.}, refining “person” with attributes to “the woman in a blue upper garment in frame 3”). Meanwhile, it made minimal grammatical adjustments (\textit{e.g.}, quantifiers, prepositions or verbs) only when necessary, thereby preserving the structural integrity of the majority of queries.

\paragraph{Question-Answering Annotation} We performed Question-Answering Annotation to assign ground-truth answers and simultaneously filter out low-quality QA pairs. This process involved four specialized annotators and totaled 40 person-days of effort at a cost of \$1,000. In the first round, two specialists divided the dataset (containing over 7,000 visual samples) equally and independently annotated their respective halves, producing initial labels for each VQA pair. During this round, they explicitly discarded VQA pairs with ambiguous referents, unnatural phrasing, or missing correct options, retaining only well-formed and verifiable pairs (corresponding to 5,394 final visual samples) for subsequent verification. In the second round, one senior expert independently re-annotated every VQA pair and compared the result with the first-round label. If the labels matched, the pair was tentatively accepted; if they disagreed, the expert re-evaluated the pair to adjudicate the discrepancy, either upholding the initial label or correcting it. In the third round, VQA pairs with consistent labels from the previous two rounds underwent a rapid check, while significant attention was devoted to verifying those pairs with prior disagreements. This intensive cross-validation ensured the precision of the final ground-truth answers.

\paragraph{Dataset Assembly} 
As is common in real-world scenarios, logistics safety surveillance data also exhibited a long-tail distribution, where the occurrences of core objects and safety hazards were highly uneven. To achieve a trade-off between diversity and balance, we rebalanced the VQA pairs across multiple dimensions, such as hallucination proportions, core object types and hazard categories, ultimately yielding \ourbench.

\begin{figure*}[htbp]  
    \centering
    \includegraphics[width=0.95\textwidth]{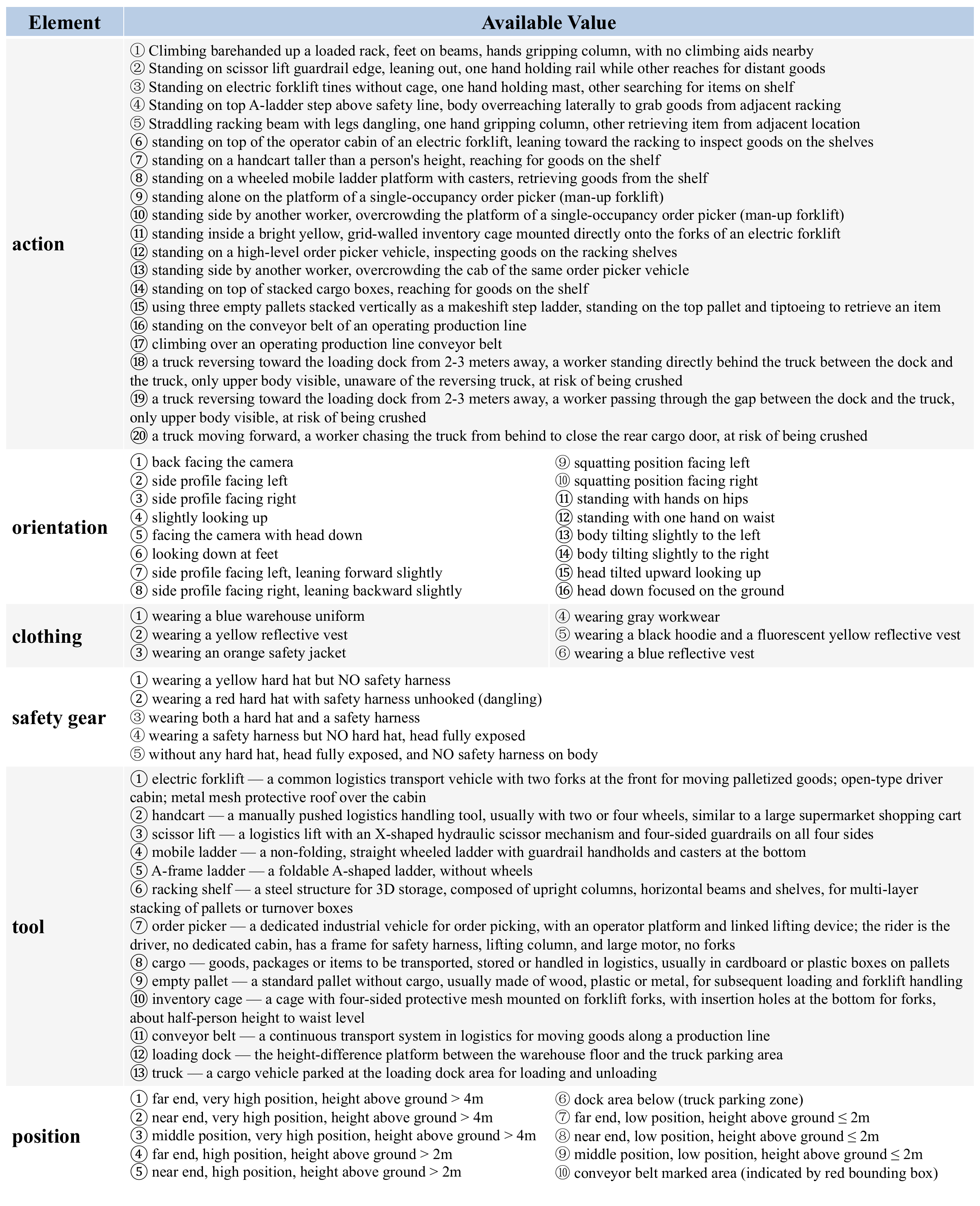}
    \caption{Values of elements for targeted image editing.}
    \label{fig:gen_prompt_value}
\end{figure*}

\begin{figure*}[htbp]  
    \centering
    \includegraphics[width=0.85\textwidth]{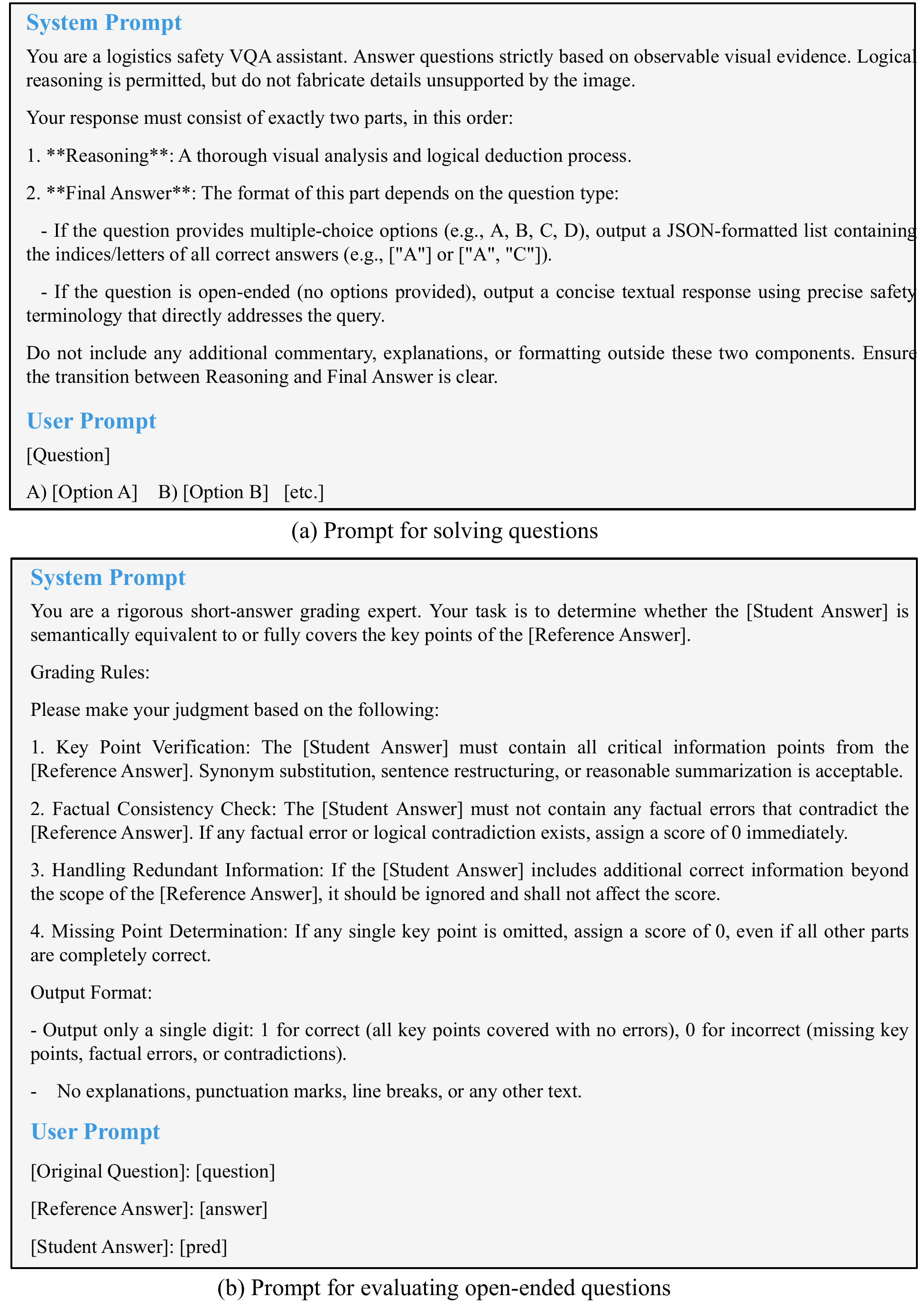}
    \caption{Prompts for solving and evaluating questions.}
    \label{fig:eval_prompt}
\end{figure*}

\subsection{Detailed Risk Definitions}
We provide precise definitions for 20 specific risks in Fig.~\ref{fig:risk_definition}, which are involved in the potential risk reasoning tasks.

\subsection{Full Task Taxonomy}
\label{task_taxonomy}
Fig.~\ref{fig:task_taxonomy} presents the full task taxonomy.

\subsection{Prompt for Targeted Image Editing}
\label{image_editing}
As illustrated in Fig.~\ref{fig:gen_prompt_template}, our prompt template for targeted image editing employs multiple placeholders, accompanied by a concrete example demonstrating its usage. The diversity of resulting prompts stems from the rich set of allowable values for each placeholder, which we comprehensively enumerate in Fig.~\ref{fig:gen_prompt_value}.

\begin{table*}[hb]
  \centering
  \caption{Stability and human alignment of LLM-as-a-Judge evaluation on 100 VQA pairs.}
  \label{tab:judge_stability}
  \begin{tabular}{ccccccc}
    \toprule
    Dimension & Metric & PAC & FM & PSD & EOC & Overall \\
    \midrule
    Intra-judge & Agreement rate & 92.9\% & 95.1\% & 93.8\% & 91.6\% & \textbf{93.3\%} \\
    Inter-judge & Agreement rate & 85.3\% & 89.4\% & 87.9\% & 85.1\% & \textbf{87.2\%} \\
    \bottomrule
  \end{tabular}
\end{table*}

\begin{table*}[ht]
\centering
\caption{Inference configurations for proprietary (purple) and open-source (green) LMMs.}
\label{tab:inference_config}

\tcbset{
  apiCard/.style={
    colback=white,
    colframe=blue!60!black,    
    colbacktitle=blue!10!white, 
    coltitle=black,
    boxrule=0.8pt,
    arc=1mm,
    width=\textwidth,
    left=6pt, right=6pt, top=1pt, bottom=1pt,
    fonttitle=\bfseries\sffamily\footnotesize,
    title=#1,
    after skip=0.02cm
  }
}

\tcbset{
  openCard/.style={
    colback=white,
    colframe=green!60!black,   
    colbacktitle=green!10!white,
    coltitle=black,
    boxrule=0.8pt,
    arc=1mm,
    width=\textwidth,
    left=6pt, right=6pt, top=1pt, bottom=1pt,
    fonttitle=\bfseries\sffamily\footnotesize,
    title=#1,
    after skip=0.02cm
  }
}

\newcommand{\sep}{\,\textbar\,}


\begin{tcolorbox}[apiCard={GPT-5.4 \& GPT-5.5}]
  \footnotesize
  \textbf{Link:} \url{https://openai.com/index/introducing-gpt-5-5/} \\
  \textbf{Temperature:} $0.2$ \sep \textbf{Max new tokens (think/non-think):} $4096$ / $256$ \sep \textbf{Eval date:} June 30-July 15, 2026
\end{tcolorbox}

\begin{tcolorbox}[apiCard={Claude-Sonnet-4.6 \& Opus-4.7}]
  \footnotesize
  \textbf{Sonnet Access:} \url{https://www.anthropic.com/claude/sonnet} \\
  \textbf{Opus Access:} \url{https://www.anthropic.com/claude/opus} \\
  \textbf{Temperature:} $0.2$ \sep \textbf{Max new tokens (think/non-think):} $4096$ / $256$ \sep \textbf{Eval date:} June 30-July 15, 2026
\end{tcolorbox}

\begin{tcolorbox}[apiCard={Gemini-3.1-Pro}]
  \footnotesize
  \textbf{Link:} \url{https://deepmind.google/models/model-cards/gemini-3-1-pro/} \\
  \textbf{Temperature:} $0.2$ \sep \textbf{Max new tokens (think/non-think):} $4096$ / $256$ \sep \textbf{Eval date:} June 30-July 15, 2026
\end{tcolorbox}

\begin{tcolorbox}[apiCard={Qwen3.7-Plus}]
  \footnotesize
  \textbf{Link:} \url{https://qwen.ai/blog?id=qwen3.7-plus} \\
  \textbf{Temperature:} $0.2$ \sep \textbf{Max new tokens (think/non-think):} $4096$ / $256$ \sep \textbf{Eval date:} June 30-July 22, 2026
\end{tcolorbox}

\vspace{0.1cm} 


\begin{tcolorbox}[openCard={GLM-5.2 \& GLM-4.7}]
  \footnotesize
  \textbf{Link:} \url{https://huggingface.co/zai-org/GLM-5.2} \\
  \textbf{Max len:} $32k$ \sep \textbf{Batch size:} $1$ \sep \textbf{backend:} Pytorch \\
  \textbf{Temperature:} $0.2$ \sep \textbf{Max new tokens (think/non-think):} $4096$ / $256$ \sep \textbf{Eval date:} July 5-22, 2026
\end{tcolorbox}

\begin{tcolorbox}[openCard={MiMo-VL-7B-RL}]
  \footnotesize
  \textbf{Link:} \url{https://huggingface.co/XiaomiMiMo/MiMo-VL-7B-RL} \\
  \textbf{Max len:} $32k$ \sep \textbf{Batch size:} $1$ \sep \textbf{backend:} Pytorch \\
  \textbf{Temperature:} $0.2$ \sep \textbf{Max new tokens (think/non-think):} $4096$ / $256$ \sep \textbf{Eval date:} July 10-15, 2026
\end{tcolorbox}

\begin{tcolorbox}[openCard={Kimi-K2-Thinking \& Kimi-K2.6}]
  \footnotesize
  \textbf{Link:} \url{https://huggingface.co/moonshotai/Kimi-K2.6} \\
  \textbf{Max len:} $32k$ \sep \textbf{Batch size:} $1$ \sep \textbf{backend:} Pytorch \\
  \textbf{Temperature:} $0.2$ \sep \textbf{Max new tokens (think/non-think):} $4096$ / $256$ \sep \textbf{Eval date:} June 25-July 22, 2026
\end{tcolorbox}

\begin{tcolorbox}[openCard={InternVL-3.5}]
  \footnotesize
  \textbf{Link:} \url{https://huggingface.co/OpenGVLab/InternVL3_5-8B} \\
  \textbf{Max len:} $32k$ \sep \textbf{Batch size:} $1$ \sep \textbf{backend:} Pytorch \\
  \textbf{Temperature:} $0.2$ \sep \textbf{Max new tokens (think/non-think):} $4096$ / $256$ \sep \textbf{Eval date:} July 10-20, 2026
\end{tcolorbox}

\begin{tcolorbox}[openCard={LLaVA-v1.6-7B}]
  \footnotesize
  \textbf{Link:} \url{https://huggingface.co/llava-hf/llava-v1.6-mistral-7b-hf} \\
  \textbf{Max len:} $32k$ \sep \textbf{Batch size:} $1$ \sep \textbf{backend:} Pytorch \\
  \textbf{Temperature:} $0.2$ \sep \textbf{Max new tokens (think/non-think):} $4096$ / $256$ \sep \textbf{Eval date:} July 10-20, 2026
\end{tcolorbox}

\begin{tcolorbox}[openCard={Qwen Open-Source Suite}]
  \footnotesize
  \textbf{Models:} \textit{Qwen3.5-Plus, Qwen3.6-Flash, Qwen3-VL (8B, 30B-A3B, 32B, 235B-A22B, Plus)} \\
  \textbf{Qwen3 Access:} \url{https://huggingface.co/collections/Qwen/qwen3} \\
  \textbf{Qwen3.5 Access:} \url{https://huggingface.co/collections/Qwen/qwen35} \\
  \textbf{Qwen3.6 Access:} \url{https://huggingface.co/collections/Qwen/qwen36} \\
  \textbf{Max len:} $32k$ \sep \textbf{Batch size:} $1$ \sep \textbf{backend:} Pytorch \\
  \textbf{Temperature:} $0.2$ \sep \textbf{Max new tokens (think/non-think):} $4096$ / $256$ \sep \textbf{Eval date:} June 25-July 22, 2026
\end{tcolorbox}

\end{table*}

\section{Evaluation Details}
\subsection{More Decoding Settings}
\label{app:decoding_setups}
Fig.~\ref{fig:eval_prompt} illustrates two evaluation prompt templates. The left one is designed for solving multi-choices and open-ended questions, while the right one is utilized in the LLM-as-a-Judge workflow for a third-party LMM to assess the correctness of the model's responses (outputting ``correct'' or ``incorrect''). In this judging workflow, the ground truth for short-answer questions typically comprises several core key points. A response is judged as ``correct'' if it contains all these essential elements, regardless of any additional elaboration; conversely, the omission of any single key point results in an ``incorrect'' classification. This zero-tolerance policy for omissions is designed to prioritize high recall in hazard detection, given the severe consequences of missed risks in industrial environments. Such a rigorous criterion serves as a stringent test, effectively validating a model's true capability and reliability in logistics safety surveillance. 
For improved transparency and reproducibility, Tab.~\ref{tab:inference_config} includes links to official documentation or model checkpoints of the evaluated LMMs, along with detailed inference configurations and the evaluation timeframe.

\subsection{Human Evaluation Protocol}
\label{app:human_eval_protocol}
To obtain comprehensive human performance, we recruited two distinct groups of participants with contrasting expertise levels. The first group consisted of 10 undergraduate student volunteers who served as generalist evaluators. While possessing strong general common sense and basic visual recognition capabilities, they lacked specialized knowledge in logistics operations or safety protocols. These students collectively covered the entire benchmark by answering approximately 1,000 questions each over a period of two weeks, receiving a stipend of \$15 per person. The second group comprised a single professional domain expert with at least four years of hands-on experience in logistics safety and surveillance. Unlike the generalists, this expert leveraged deep domain knowledge (\textit{e.g.}, familiarity with industrial equipment and safety protocols) to independently answer all questions in the benchmark. This expert evaluation also spanned two weeks and incurred a total cost of \$700. The results labeled “Novice Student” and “Logistics Expert” in Tab.~\ref{tab:main_res} correspond to the performance of the student volunteers and logistics expert, respectively.

\subsection{Reliability of LLM-as-Judge Evaluation}
A reliable LLM-as-Judge evaluation must be both stable and aligned with human expert judgments. We therefore randomly sampled 100 questions from the open-ended VQA set and examined the judge along the following two dimensions: 
\begin{itemize}
    \item Intra-judge: The same LLM judge evaluates each question 3× with temperature = 0.7. We report the agreement rate across the three runs.
    \item Inter-judge: Human experts independently judge the same subset. We report the agreement rate between the LLM judge's decisions and the human judgments.
\end{itemize}

As illustrated in Tab.~\ref{tab:judge_stability}, intra-judge repeatability exceeds 93\% and the LLM judge's agreement with human experts exceeds 87\%. These results prove that the LLM judge is stable and aligned with human judgment.

\section{Visualization}
\begin{figure*}[htbp]  
    \centering
    \includegraphics[width=1.0\textwidth]{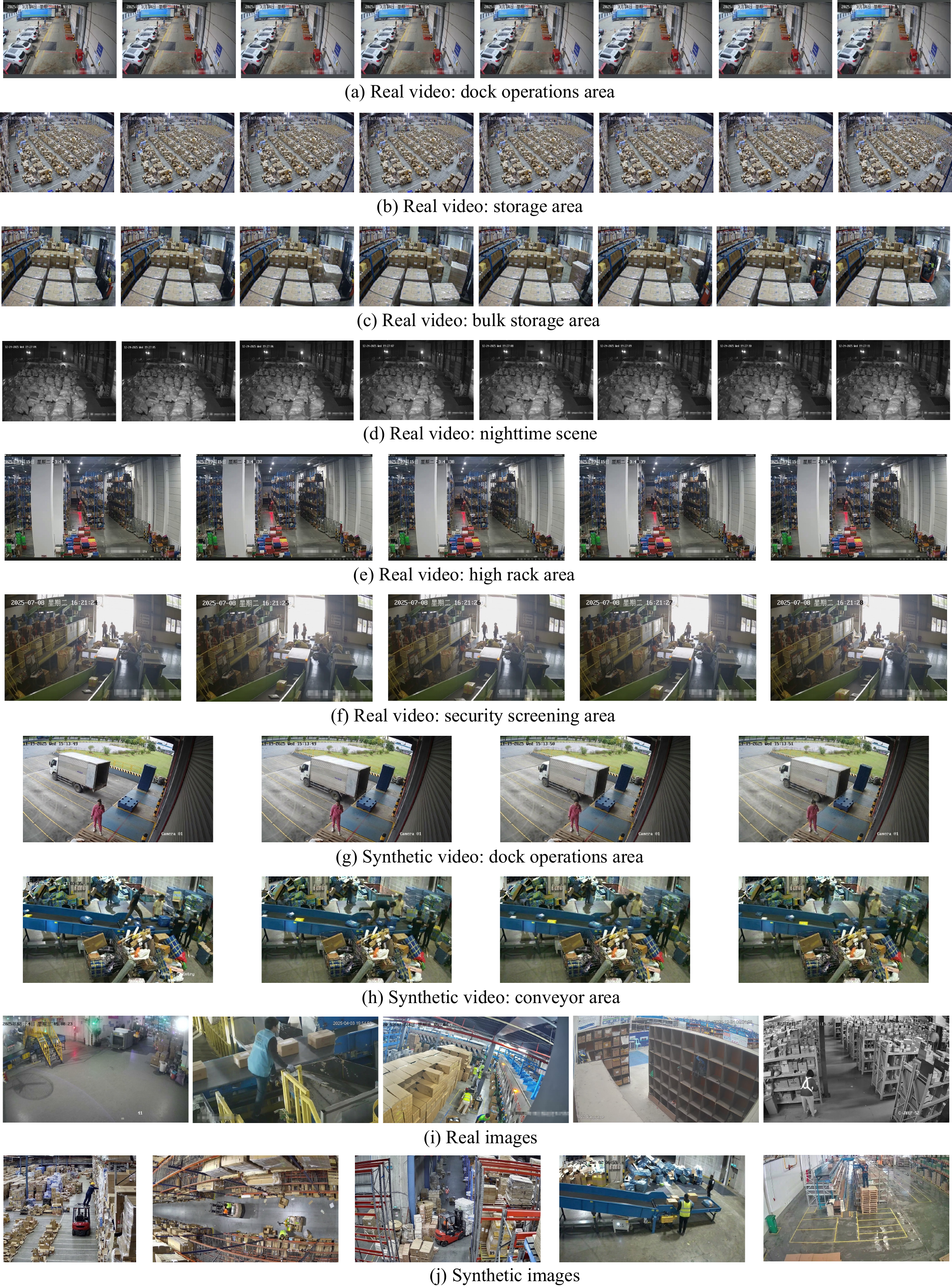}
    \caption{Examples of real and synthetic visual data, covering both daytime and nighttime scenes across various zones within logistics warehouses.}
    \label{fig:visual_vis}
\end{figure*}

To facilitate a more intuitive understanding, we provide additional visualization examples. Fig.~\ref{fig:visual_vis} displays visual-only examples (raw images).

\end{document}